\documentclass[letterpaper,10pt,journal,twoside]{IEEEtran}
\usepackage{listings}
\usepackage[auth-lg]{authblk}
\usepackage{algpseudocode}                                 % new algorithm package
\usepackage{algorithm}
\usepackage{graphicx}                                      % include pdf figures
\usepackage{amsmath}
\usepackage{amssymb}
\usepackage{amsfonts}
\usepackage{booktabs}
\usepackage{multirow}
\usepackage[subrefformat=parens,farskip=0pt,justification=centering]{subfig}
\usepackage{color}
\usepackage{cite}                                          % more citations in one bracket
\usepackage{comment}                                       % use comment
\usepackage{soul}                                          % use highlight command \hl{}
\soulregister\cite7
\soulregister\ref7
\soulregister\pageref7
\usepackage{amsthm}
\usepackage{etoolbox}                                      % commands \newtoggle, \toggletrue, \iftoggle
\usepackage{url}
\usepackage{hyperref}
\usepackage{nth}                                           % nth command
\usepackage{bm}                                            % bm command
\usepackage{balance}
\usepackage{threeparttable}
\usepackage{booktabs}
\usepackage{colortbl}
\usepackage{filecontents}                                  % support to pgfplots
\usepackage{pgfplots}
\usepackage{pgfplotstable}
\pgfplotsset{compat=newest}
\usepackage{caption}
\usepackage[normalem]{ulem}                                            % for underline
\usepackage{mathrsfs}
\usepackage{subeqnarray}

\usepackage{enumitem}

\algrenewcommand\textproc{\texttt}
\newcommand{\tabincell}[2]{
    \begin{tabular}{@{}#1@{}}#2\end{tabular}
}

\makeatletter
\let\OldStatex\Statex
\renewcommand{\Statex}[1][3]{%
  \setlength\@tempdima{\algorithmicindent}%
  \OldStatex\hskip\dimexpr#1\@tempdima\relax
}
\makeatother

\RequirePackage[normalem]{ulem} %DIF PREAMBLE
\RequirePackage{color}\definecolor{RED}{rgb}{1,0,0}\definecolor{BLUE}{rgb}{0,0,1} %DIF PREAMBLE
\graphicspath{{./}}

\lstdefinestyle{base}
{
    language   = C,
    emptylines = 1,
    breaklines = true,
    basicstyle = \ttfamily\scriptsize,
    moredelim  = **[is][\color{red}]{@}{@},
}

\begin{document}

\title{Efficient Computation Reduction in Bayesian Neural Networks through Feature Decomposition and Memorization}

\author{Xiaotao~Jia,~\IEEEmembership{Member,~IEEE,}
        Jianlei~Yang,~\IEEEmembership{Member,~IEEE,}
        Runze~Liu,
        Xueyan~Wang,~\IEEEmembership{Member,~IEEE,}
        Sorin~Dan~Cotofana,~\IEEEmembership{Fellow,~IEEE}
        and~Weisheng~Zhao,~\IEEEmembership{Fellow,~IEEE}
\thanks{Manuscript received on June 2019; revised on December 2019; accepted on April 2020. This work was supported in part by the National Natural Science Foundation of China (61602022, 61701013), State Key Laboratory of Software Development Environment (SKLSDE-2018ZX-07), National Key Technology Program of China (2017ZX01032101), CCF-Tencent IAGR20180101, State Key Laboratory of Computer Architecture (CARCH201917) and the 111 Talent Program B16001. \textit{Corresponding authors are Jianlei Yang and Weisheng Zhao.}}
\thanks{X. Jia, X. Wang and W. Zhao are with Fert Beijing Research Institute, School of Microelectronics, BDBC, Qingdao Research Institute, Beihang University, Beijing, 100191, China. E-mail: weisheng.zhao@buaa.edu.cn}
\thanks{J. Yang and R. Liu are with Fert Beijing Research Institute, School of Computer Science and Engineering, BDBC, Beihang University, Beijing, 100191, China. E-mail: jianlei@buaa.edu.cn}
\thanks{S. D. Cotofana is with the Electrical Engineering, Mathematics and Computer Science Faculty, Delft University of Technology, Delft, the Netherlands.}
}

\maketitle
\thispagestyle{empty}

\begin{abstract}
Bayesian method is capable of capturing real world uncertainties/incompleteness and properly addressing the over-fitting issue faced by deep neural networks. In recent years, Bayesian Neural Networks (BNNs)  have drawn tremendous attentions of AI researchers and proved to be successful in many applications. However, the required high computation complexity makes BNNs difficult to be deployed in computing systems with limited power budget.
In this paper, an efficient BNN inference flow is proposed to reduce the computation cost then is evaluated by means of both software and hardware implementations. A feature decomposition and memorization (\texttt{DM}) strategy is utilized to reform the BNN inference flow in a reduced manner. About half of the computations could be eliminated compared to the traditional approach that has been proved by theoretical analysis and software validations. Subsequently, in order to resolve the hardware resource limitations, a memory-friendly computing framework is further deployed to reduce the memory overhead introduced by \texttt{DM} strategy. Finally, we implement our approach in Verilog and synthesise it with 45 $nm$ FreePDK technology. Hardware simulation results on multi-layer BNNs demonstrate that, when compared with the traditional BNN inference method, it provides an energy consumption reduction of 73\% and a 4$\times$ speedup at the expense of 14\% area overhead.

\end{abstract}


\IEEEpeerreviewmaketitle

 \iffalse

 \bibliography{../ref/BNN}

 \fi

\section{Introduction} \label{Section:Introduction}

Deep Learning paradigm has created the premises for the development of several Deep Neural Network (DNN) models \cite{lecun1995convolutional,hochreiter1997long,goodfellow2014generative}. DNNs have been utilized as underlying implementation tools to boost various applications, e.g., computer vision~\cite{krizhevsky2012imagenet}, natural language processing~\cite{sutskever2014sequence}, speech recognition~\cite{hinton2012deep}, autonomous driving~\cite{chen2015deepdriving}. They made DNN research as one of the most active Artificial Intelligence (AI) branches. However, even though practical utilization of DNNs provides very promising results, e.g., in some cases DNNs could even outperform human capabilities \cite{he2016deep}, their global proliferation is mainly impeded by the lack of proper theoretical justification. DNN training is an optimization procedure which relies on Maximum Likelihood Estimation (MLE) of the synaptic weights. However, it is well understood and accepted that MLE is not able to properly handle the inherent weights uncertainties. This implies that, from a practical standpoint, MLE based training is susceptible to over-fitting, as experimentally observed in many DNN behaviours~\cite{bishop2006pattern}. Furthermore, DNNs have other disadvantages such as data-hungry, lack of solid mathematical foundations, and easy to be fooled~\cite{nguyen2015deep}.

To address these shortcomings, Bayesian methods have been introduced by providing mathematically grounded approaches. Bayesian methods could achieve reasonable learning accuracy from small datasets and exhibit the required robustness to address the over-fitting issue~\cite{gal2015bayesian}. More importantly, they could inherently deal with real world uncertainty and take prior knowledge into consideration. Thus, the potential combination of the complimentary strengths of Bayesian methods and DNNs is receiving increasing interest and Bayesian Neural Networks (BNNs) have been applied in many different applications~\cite{ticknor2013bayesian,spinbis,jia2017spintronics,chien2016bayesian}.  Several probabilistic programming frameworks have been developed, such as Edward~\cite{tran2017edward}, Pyro~\cite{bingham2018pyro} and Zhusuan~\cite{shi2017zhusuan}, which could provide efficient implementations for Bayesian Deep Learning (BDL).

However, the remarkable DNN capabilities can only be exploited at the expense of a high computation complexity associated to the deep layer structure, which requires the utilization of high performance computing resources to accelerate the training and inference procedures. To resolve this, numerous approaches have been proposed to reduce the computation cost by means of, e.g., network pruning~\cite{han2015learning}, low rank approximation~\cite{denton2014exploiting}, and structure sparsity learning~\cite{wen2016learning}.  However, those have rather limited impact as DNN models have been wildly applied in Internet-of-Things (IoTs) and embedded systems, which are usually computation resource and power consumption confined. Aiming to enable the DNN technology on such devices, the network inference tasks are usually performed on highly-parallel and specialized ASIC hardware~\cite{knag2015sparse}. As for BNN, an FPGA-based accelerator has been recently proposed in~\cite{cai2018vibnn}.

In view of the above, a novel BNN inference approach which could reduce computation complexity is proposed in this paper. It relies on a new mathematical formulation that enables computation reuse and consequently has a significant impact on computation latency and energy consumption. This work only focuses on the inference part as we can assume that the training is not done in-place for IoT applications (actually Edward framework is adopted for BNN training in this work), thus it does not affect the available power and computation resources budget. The main contributions of the paper are summarized as follows:
\begin{itemize}
    \item We introduce a feature Decomposition and Memorization (\texttt{DM}) approach applicable for BNNs whose parameters obey Gaussian distribution. The DM strategy takes advantage of single-layer BNN equations characteristics to perform fast and energy effective inference and theoretically eliminates nearly half of the traditionally required computations.
    \item In order to apply \texttt{DM} strategy into multi-layer BNNs, two methods are further introduced which are regarded as Hybrid-BNN and \texttt{DM}-BNN, respectively. Both of them could achieve the reasonable accuracy with less energy and less runtime.
    \item A memory friendly computing framework is proposed to mitigate the \texttt{DM} associated memory overhead without any detrimental effect on the inference computation cost and result quality.
    \item The proposed approaches and the traditional BNN inference have been evaluated by means of both software and hardware implementations. The software implementation indicates that the proposed \texttt{DM}-BNN could reduce the computation cost by 85\% at the cost of very slight accuracy loss. The hardware implementation suggests that \texttt{DM}-BNN reduces the energy consumption by 73\% and achieves a 4$\times$ speedup at the expense of 14\% area overhead and a minor accuracy degradation.
\end{itemize}

The rest of this paper is organized as follows. Section~\ref{Section:Preliminary} discusses some preliminaries and related works. Section~\ref{Section:computation} demonstrates the proposed \texttt{DM} strategy. The memory friendly computing framework is presented in Section~\ref{Section:memory}. Experimental results are illustrated in Section~\ref{Section:experiment}. Concluding remarks are provided in Section~\ref{Section:conclusion}.

 \iffalse

 \bibliography{../ref/BNN}

 \fi

\section {Preliminaries} \label{Section:Preliminary}

In this section, BNN background is reviewed and some related works are briefly discussed.

\begin{figure}[tb!]
    \centering
    \includegraphics[width=0.45\textwidth]{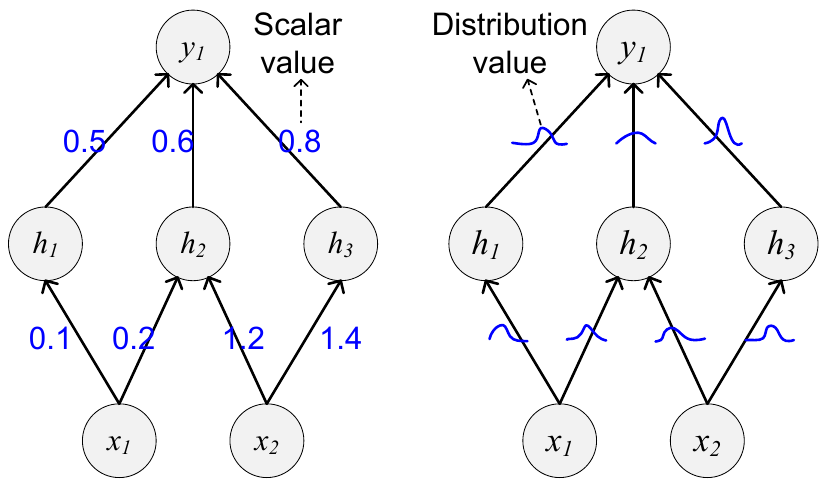}
    \caption{Different from deterministic Neural Networks (left) which have scalar weight values, BNNs (right) have the weight values defined by distributions.}
    \label{fig:bnn}
\end{figure}

By introducing the augmentation of standard neural networks with posterior inference, BNNs aims to create a deep learning framework which is able to cope with parameter uncertainty. Different from DNNs whose weight values are deterministic, BNNs offer a probabilistic interpretation of deep learning models by inferring weight value distributions, as graphically indicated in Fig.~\ref{fig:bnn}. BNNs place a prior distribution over each neural network's parameter, and the likelihood (\textit{i.e.}, the training data) is then fed into the network aiming to find the optimal posterior distribution. Usually, BNN posterior distributions fit the Gaussian profile.

To build a BNN model, we denote the BNN by the   $f^{\mathcal{W}}(\cdot)$ function, where $f$ represents the network structure and $\mathcal{W}$ is the distribution set of model parameters, \textit{i.e.}, synaptic weights and biases. 
Given a set of training datasets $\bm{X} = \{X_1, X_2, \cdots, X_N\}$ and the associated labels $\bm{y} = \{y_1, y_2, \cdots, y_N\}$, the BNN training aims to find the posterior distribution over model parameters $p(\mathcal{W}|\bm{X},\bm{y})$, which in practice  is very difficult to find since the posterior weight distribution is highly complex. Modern BNN research is mainly focused on variational inference methods~\cite{graves2011practical,blundell2015weight} or Markov chain Monte Carlo approaches~\cite{chen2014stochastic, balan2015bayesian}. In this work, we mainly focus on the prediction/inference procedure rather than training  procedure. And the Edward framework~\cite{tran2017edward} is utilized for software evaluations, which relies on a variational inference method.

BNN prediction procedure starts from instantiating a series of concrete neural networks for forward propagation. The instantiatation requires Gaussian random variable sampling for all the posterior distributions which have been identified during the training process. These sampling operations could be implemented both in software and hardware approach~\cite{malik2016gaussian}. Assuming that the posterior distribution of one weight $\omega$ fits a Gaussian distribution with location (or mean value) of $\mu$ and scale (or standard deviation) of $\sigma$, \textit{i.e.}, $\omega\ \sim N(\mu, \sigma^2)$, sampling the weight $w$ requires to select a random number $h$ from the standard Gaussian distribution $N(0,1)$. Based on the theory that if $U \sim N(0,1)$ then $X = U \sigma + \mu \sim N(\mu, \sigma^2)$, the resulted random weight is calculated as $w=\sigma h + \mu$ by the scale-location transformation. Once all of the random weights have been sampled, the BNN prediction follows the evaluation paradigm detailed in Section~\ref{Section:computation}. We notice that standard Gaussian random number generation algorithms are classified as inversion, transformation, rejection, and recursion methods~\cite{thomas2007gaussian} among which central limit theorem based transformation method is most widely used.

\begin{figure*}[tb!]
    \centering
    \includegraphics[width=0.88\textwidth]{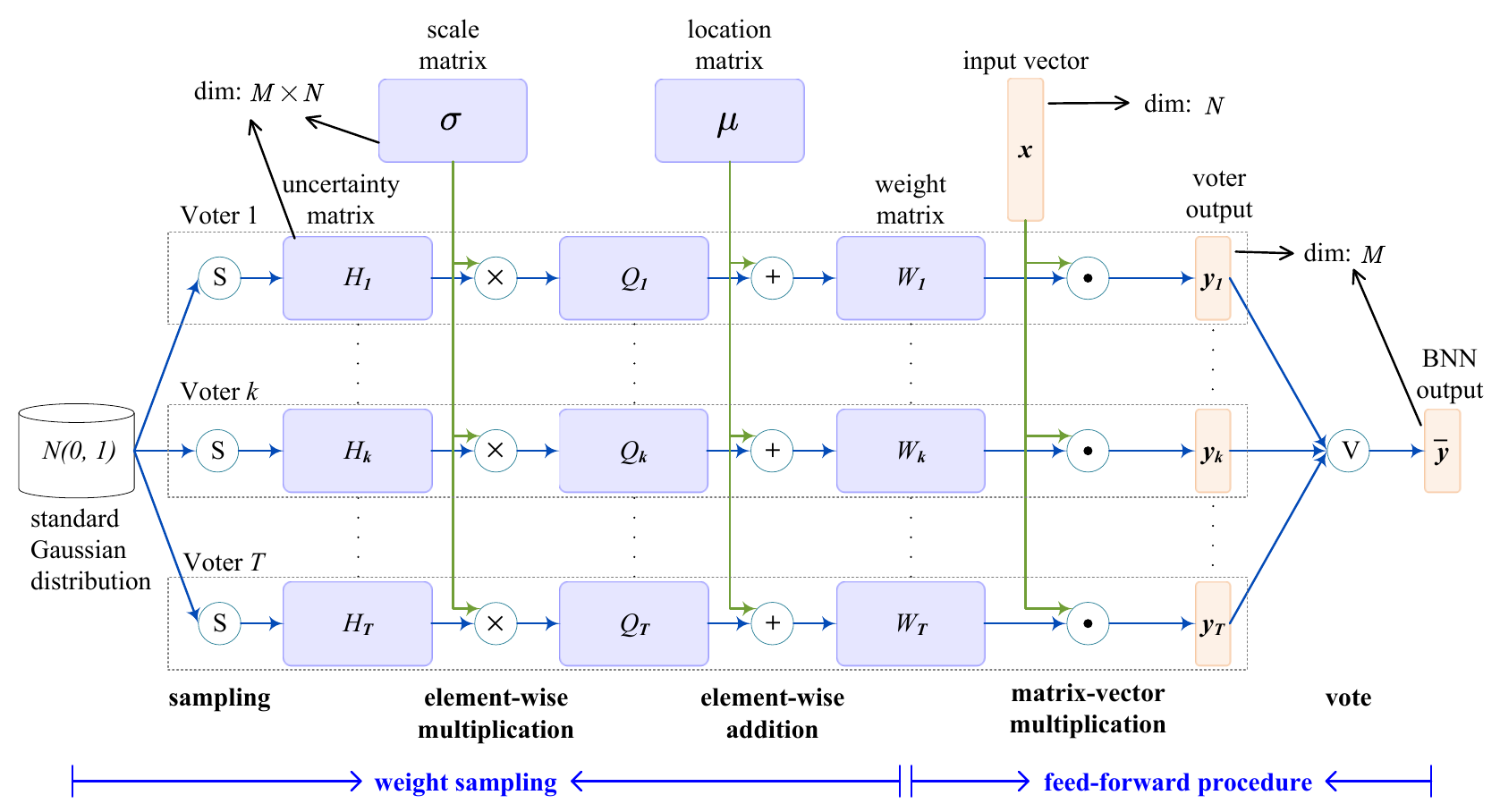}
    \caption{Single-layer BNN dataflow divided in two steps: (i) Gaussian random number generation and (ii) feed-forward propagation. For the sake of convenience, the sampling procedure is denoted as \textcircled{\small{S}} and voters evaluation is denoted as \textcircled{\small{V}}. $\sigma$ and $\mu$ are well-trained BNN weight, and $\bm{x}$ the input vector. The bias terms are not taken into account for simplify.}
    \label{fig:df-naive}
\end{figure*}

As aforementioned, there have been many research works focused on DNN hardware acceleration while very little has been reported in relation to BNN counterparts with the notable exception of VIBNN~\cite{cai2018vibnn}. The BNN prediction procedure is accelerated on FPGA platform in VIBNN by introducing two novel Gaussian random number generators, memory optimization techniques, and a deep pipeline structure. In~\cite{cai2018vibnn}, concrete neural networks are firstly instantiated based on weights distribution; then VIBNN performs DNN evaluations repeatedly on them. However, these concrete neural networks are not independent so that VIBNN could not exactly capture the nature of the BNN paradigm. In this work, a novel BNN inference framework is proposed to reduce the computation complexity and BNN-oriented architecture is implemented to improve the hardware efficiency.
 \iffalse

 \bibliography{../ref/BNN}

 \fi

\section {BNN Computation Reduction} \label{Section:computation}
% Energy and latency efficient BNN
% computation friendly
% memory friendly

In this section, a novel BNN inference method is demonstrated by taking the advantages of a fact that a certain amount of computations could be shared between its associated neural network instances. We first analyze the single-layer BNN dataflow and propose a feature Decomposition and Memorization (\texttt{DM}) approach to reduce the computation complexity. Subsequently, we extend \texttt{DM} to multi-layer BNNs and analyze its computation complexity and memory consumption.

\subsection{BNN Dataflow} \label{Section:bnn:dm}

Given a well-trained BNN, actual weights and biases sampling are required to instantiate a concrete neural network for inference procedure. However, several concrete neural networks with different parameter values are usually required for instantiation to fully explore the uncertainty using posterior distribution, rather one. Subsequently, the input data (\textit{e.g.}, an image) is fed into all the instantiated neural networks to obtain their predictions, and the actual response is determined by voting. The corresponding BNN dataflow is illustrated in Fig.~\ref{fig:df-naive} for a one-layer fully connected neural network. The involved variables of matrices, vectors are described in Table~\ref{table:notions} and defined operators are illustrated in Table~\ref{table:operations}.
\begin{table}[tb]
\centering
\caption{Bayesian Neural Network Notations.}
\label{table:notions}
\begin{tabular}{c|l|l}
    \specialrule{0.8pt}{0pt}{0pt}
     Symbol & Description & Dimension \\
    \hline
     $N$    &   input neural count      &  1 \\
     $M$    &   output neural count     &  1 \\
     $T$    &   the number of sampling      &  1 \\
     $\mathcal{L}$ & the number of BNN layers      &  1 \\
     $\mathcal{W}$ & weight distribution matrix & $M \times N$ \\
     $W$    & weight matrix      & $M \times N$   \\
     $H$    & uncertainty matrix      & $M \times N$   \\
     $\sigma$/$\mu$    & scale/location matrix      & $M \times N$   \\
     $\bm{x}$   & input vector         &  $N$ \\
     $\bm{y}$   & output vector         &  $M$ \\
     \specialrule{0.8pt}{0pt}{0pt}
\end{tabular}
\end{table}

\begin{table}[tb]
\centering
\caption{Bayesian Neural Network Operations.}
\label{table:operations}
\begin{tabular}{c|l}
    \specialrule{0.8pt}{0pt}{0pt}
     Symbol & Operation  \\
    \hline
     $+$    &   element-wise addition      \\
     $\times$    &   element-wise multiplication     \\
     $\cdot$    &   matrix-vector multiplication     \\
     $<>_L$ & line-wise inner product  \\
     \textcircled{\scriptsize{S}}    & random number sample          \\
     \textcircled{\scriptsize{V}}    & vote        \\
     \textcircled{\scriptsize{P}}    & pre-compute \\
     \textcircled{\scriptsize{F}}    & feed-forward         \\
     \textcircled{\scriptsize{T}}    & scale-location transformation         \\
     \specialrule{0.8pt}{0pt}{0pt}
\end{tabular}
\end{table}

\begin{algorithm}[tb!]
% \small
\caption{Standard BNN Evaluation}
\label{alg:bnn}
\begin{algorithmic}[1]
    \Require
        Well-trained BNN with weight parameters $\mu$ and $\sigma$
    \Require
        Input data \bm{$x$}
    \Ensure
        BNN output \bm{$\bar{y}$}
    \For {($k = 1; k \le T; k = k + 1$)}
        \State Sample uncertainty matrix $H_k$ from $N(0, 1)$   \label{alg:bnn:sample}
        \State $Q_k = H_k \times \sigma$ \label{alg:bnn:scale}
        \State $W_k = Q_k + \mu$  \label{alg:bnn:location}
        \State $\bm{y_k} = W_k \cdot \bm{x}$    \label{alg:bnn:dot}
    \EndFor 
    \State $\bm{\bar{y}} = \frac {\sum_{k=1}^{T}\bm{y_k}} {T}$   \label{alg:bnn:vote}
\end{algorithmic}
\end{algorithm}

The considered BNN contains $N$ input neurons and $M$ output neurons. $\mu$ and $\sigma$ are $M \times N$ dimensional BNN weight values of posterior distribution parameters, regarded as location matrix and scale matrix, respectively. $T$ is the number of NN samples to be evaluated in order to obtain the appropriate BNN's response. Based on the NN theory, an NN's output is calculated by Eqn.~\eqref{Equation:nn}, which means that when deriving BNN's output Eqn.~\eqref{Equation:nn} will be evaluated for $T$ times. 
\begin{equation}
\label{Equation:nn}
\bm{y} = W \bm{x} + \bm{b}
\end{equation}

\begin{figure*}[tb!]
    \centering
    \includegraphics[width=0.9\textwidth]{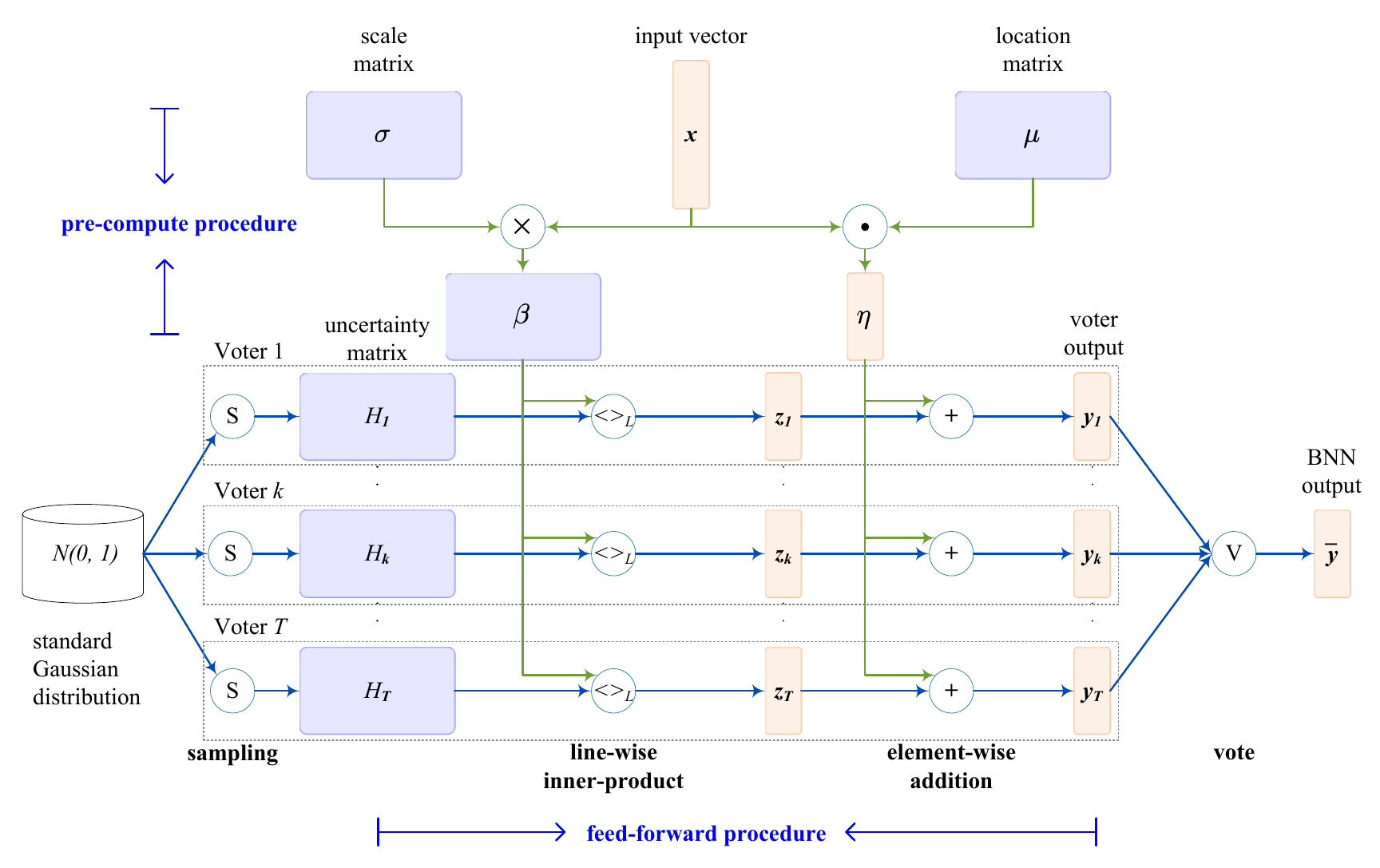}
    \caption{Single-layer BNN dataflow with feature Decomposition and Memorization. The bias terms are not considered. $\beta$ and $\eta$ is pre-computed and stored in local memory.}
    \label{fig:df-dm}
\end{figure*}

Compared with the computation cost of matrix-vector multiplication between $W$ and $x$, that of vector addition between $Wx$ and $b$ could be neglected. Hence, the bias terms are not taken into consideration in the following complexity analysis. Fig.~\ref{fig:df-naive} graphically describes the standard single-layer BNN dataflow and Algorithm~\ref{alg:bnn} details the computation procedure as follows:
\begin{enumerate}[label=(\roman*)]
    \item  $T$ concrete weight matrices $W_1, W_2, \cdots, W_T$ are sampled according to the weight posterior distributions by exploiting Gaussian Random Number Generators (GRNG) (Line~\ref{alg:bnn:sample} -- \ref{alg:bnn:location});
    \item Matrix-vector multiplication operation is performed between the input \bm{$x$} and each weight matrix to generate $T$ outputs $\bm{y_1}, \bm{y_2}, \cdots,\bm{y_T}$ (Line~\ref{alg:bnn:dot});
    \item Output result \bm{$\bar{y}$} is computed by averaging voter results of  $\bm{y_1}, \bm{y_2}, \cdots,\bm{y_T}$ (Line~\ref{alg:bnn:vote}).
\end{enumerate}

As clearly suggested by Fig.~\ref{fig:df-naive}, there are $T$ sampling dataflow which could be performed in parallel. And each of them could be regarded as a \textbf{voter} who contributes to the output results calculation in the final voting stage.

\subsection{Feature Decomposition and Memorization} \label{Section:computation:dm}

As previously discussed, the instantiated NNs are evaluated in parallel, then the following voting procedure determines the final output. And obviously improving the number of instantiations could improve the BNN model robustness but will introduce more computation cost. In this section, the BNN inference dataflow is reformulated through the proposed \texttt{DM} strategy to reduce the total computation complexity. It is based on the observations that a certain amount of computations could be actually shared among different instantiated NNs.

\begin{figure*}[tb!]
    \centering
    \includegraphics{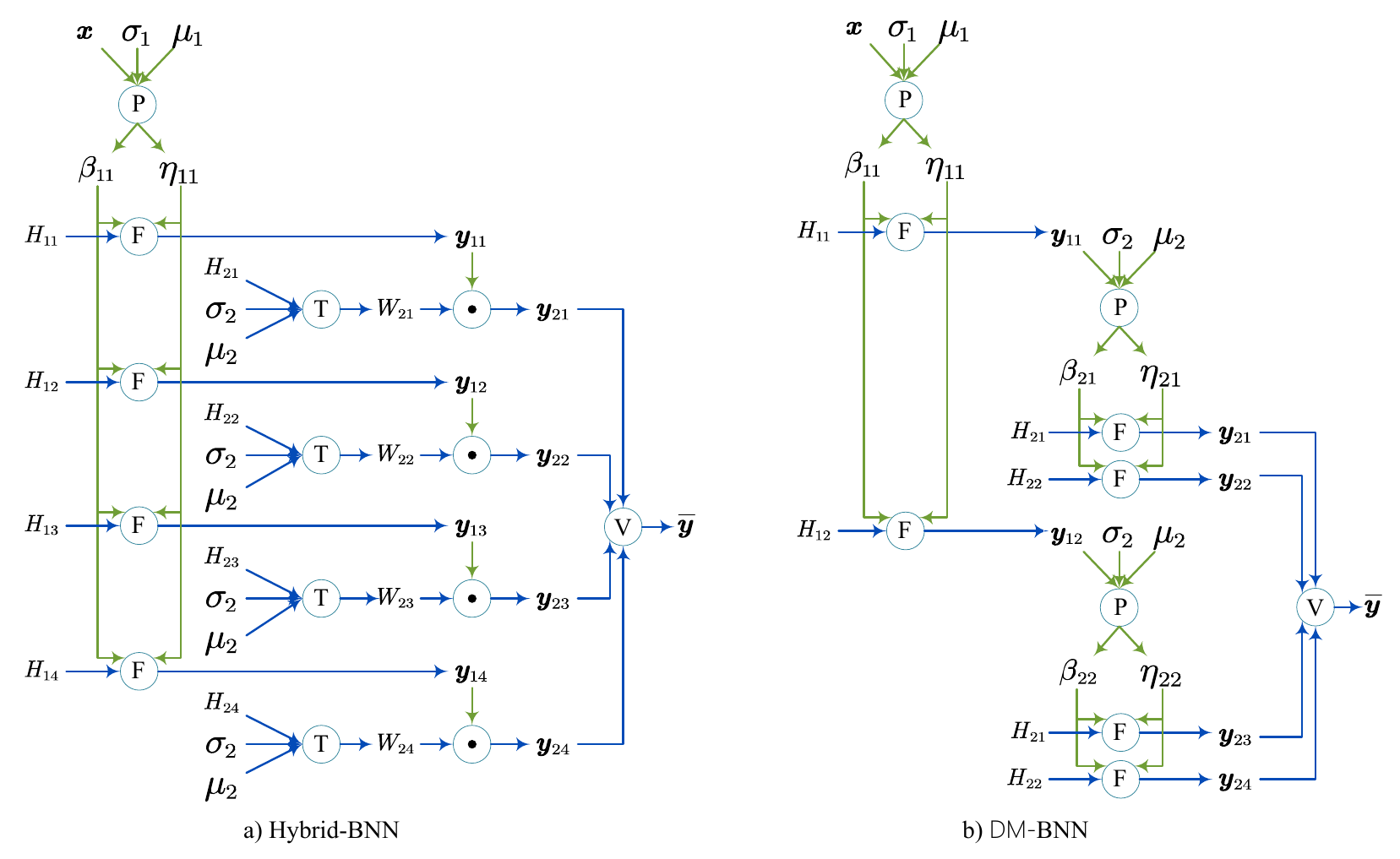}
    \caption{Simplified dataflow of multi-layer BNNs (with one hidden layer), where \textcircled{\small{P}}, \textcircled{\small{F}}, and \textcircled{\small{T}} stand for pre-compute, feed-forward, and scale-location transformation, respectively. a) Hybrid-BNN: The \texttt{DM} strategy is only applied for the first layer. b) \texttt{DM}-BNN: The \texttt{DM} strategy is applied for all layers.}
    \label{fig:deep-bnn}
\end{figure*}

\subsubsection{\texttt{DM} for single-layer BNNs}

To evaluate BNN's response for a given input, each voting result $\bm{y_k}$ is calculated by performing multiplication and addition on the input $\bm{x}$ and sampled weights $W_k$ by \eqref{Equation:nn}. Each element of $\bm{y_k}$ is calculated as follows:
\begin{subequations}
\label{Equation:bnnelement}
\begin{align}
% 3 lines
% y_k^i & = \sum_{j=1}^{N}{w_{k}^{ij}x^{j}} & \\
%       & = \sum_{j=1}^{N}{\left(h_{k}^{ij}\sigma^{ij}+\mu^{ij}\right)x^{j}} & \label{Equation:bnnelement:b}\\
%       & = \sum_{j=1}^{N}{h_{k}^{ij}\uwave{\sigma^{ij}x^{j}}}+
%           \sum_{j=1}^{N}{\uwave{\mu^{ij}x^{j}}} & \label{Equation:bnnelement:c}
% 2 lines
y_k^i & = \sum_{j=1}^{N}{w_{k}^{ij}x^{j}} = \sum_{j=1}^{N}{\left(h_{k}^{ij}\sigma^{ij}+\mu^{ij}\right)x^{j}} & \label{Equation:bnnelement:b}\\
      & = \sum_{j=1}^{N}{h_{k}^{ij}\uwave{\sigma^{ij}x^{j}}}+
          \sum_{j=1}^{N}{\uwave{\mu^{ij}x^{j}}} & \label{Equation:bnnelement:c}
\end{align}
\end{subequations} where $k = 1, 2, \cdots, T$ and $i = 1, 2, \cdots, M$, $w_{k}^{ij}$ is sampled according to $\sigma_{k}^{ij}$ and $\mu_{k}^{ij}$, $h_{k}^{ij}$ is introduced to represent the uncertainty.

Following Eqn.~\eqref{Equation:bnnelement:b} the single-layer BNN inference dataflow has been illustrated in Fig.~\ref{fig:df-naive}. Noticed that the input $\bm{x}$, and posterior distribution parameters $\sigma$ and $\mu$ are the same for different voting evaluations, we could consider to reuse these computations among them. The features expressed in Eqn.~\eqref{Equation:bnnelement:b} could be equivalently \textbf{decomposed}  as Eqn.~\eqref{Equation:bnnelement:c} in which the computation results of $\sigma \times \bm{x}$ and $\mu \cdot \bm{x}$ are the same for all the $T$ voters.  
Such potential computation sharing is the underlying property utilized in the proposed \texttt{DM} strategy, which  \textbf{memorizes} $\sigma^{ij}x^j, (i=1,2,\cdots,M; j=1,2,\cdots,N)$ and $\sum_{j=1}^{N}{\mu^{ij}}x^j, (i=1,2,\cdots,M)$
so that they could be directly loaded from memory instead of being $T \times$ recomputed during the voters evaluation stage.

\begin{algorithm}[tb!]
% \small
\caption{\texttt{DM} based BNN Evaluation}
\label{alg:dfbnn}
\begin{algorithmic}[1]
    \Require
        Well-trained BNN with weight parameters $\mu$ and $\sigma$
    \Require
        Input data \bm{$x$}
    \Ensure
        BNN output \bm{$\bar{y}$}
    \State $\eta = \mu \cdot \bm{x}$ \label{alg:dfbnn:mu_x}
    \State $\beta = \sigma \times \bm{x}$ \label{alg:dfbnn:sigma_x}
    \For {($k = 1; k \le T; k = k + 1$)} \label{alg:dfbnn:beginfor}
        \State Sample uncertainty matrix $H_k$ from $N(0, 1)$   \label{alg:dfbnn:sample}
        \State $\bm{z_k} = <H_k,\beta>_L$   \label{alg:dfbnn:h_beta}
        \State $\bm{y_k} = \bm{z_k} + \eta$    \label{alg:dfbnn:z_eta}
    \EndFor \label{alg:dfbnn:endfor}
    \State $\bm{\bar{y}} = \frac {\sum_{k=1}^{T}\bm{y_k}} {T}$
\end{algorithmic}
\end{algorithm}

Following Eqn.~\eqref{Equation:bnnelement:c} the BNN inference dataflow could be depicted as Fig.~\ref{fig:df-dm} with the proposed \texttt{DM} strategy. Compared with standard dataflow as shown in Fig.~\ref{fig:df-naive}, the multiplications of $\bm{x}$ and $\sigma$ (as well as $\bm{x}$ and $\mu$) is pre-computed and stored in local memory. Hence, the BNN evaluation could be performed directly with the \texttt{uncertainty matrices} $H_k$ (sampled from standard Gaussian distribution) without scale-location transformation. For each BNN evaluation, the voter's response $\bm{y_k}$ could be obtained according to the pre-computed features and sampled uncertainty matrix $H_k$. The BNN inference dataflow with \texttt{DM} strategy is described in Algorithm~\ref{alg:dfbnn}. 
% The so-called \texttt{DM} strategy is defined by pre-computing $\eta$ and $\beta$ (Lines~\ref{alg:dfbnn:mu_x}-\ref{alg:dfbnn:sigma_x}) which could be regarded as feature decomposition. 
% In Line~\ref{alg:dfbnn:mu_x}, `$\cdot$' means matrix-vector product operation. 
In Line~\ref{alg:dfbnn:sigma_x}, `$\times$' means that vector $\bm{x}$ performs element-wise product with every row of $\sigma$, which implies that $\beta$ has the same dimension as $\sigma$. And obviously the resulted $\eta$ and $\beta$ requires additional memory space to store them. In line~\ref{alg:dfbnn:h_beta}, the operation `$<>_L$' indicates line-wise inner product operation, \textit{i.e.}, an inner-product is performed on each row of $W$ and $\beta$. In Algorithm~\ref{alg:dfbnn}, Lines~\ref{alg:dfbnn:mu_x}-\ref{alg:dfbnn:sigma_x} and Lines~\ref{alg:dfbnn:beginfor}-~\ref{alg:dfbnn:endfor} correspond to the pre-compute and feed-forward stage in Fig.~\ref{fig:df-dm}, respectively. For the sake of convenience, the pre-compute stage is denoted as \textcircled{\scriptsize{P}} and feed-forward evaluation is denoted as \textcircled{\scriptsize{F}}.  

\subsubsection{\texttt{DM} for multi-layer BNNs}\label{Section:bnn:dbnn}

As it is well known that Deep Neural Networks exhibit more intrinsic computation capabilities than single-layer counterparts when dealing with complex input-output mapping. It is of obvious interest to extend the \texttt{DM} strategy for multi-layer Bayesian networks. As shown in Fig.~\ref{fig:df-dm}, single-layer network has a $1$-to-$T$ relationship between input vector and output vector which means that all voters receive the same input $\bm{x}$. It is the key condition behind the DM strategy utilization. As for multi-layer BNN, we can get $T$ output vectors after the computation of the first layer which are also the input vectors of the second layer. That is to say, there is a $T$-to-$T$ relationship between the input vector and output vector rather than $1$-to-$T$. As a result the proposed \texttt{DM} strategy cannot be directly applied in all layers of multi-layer BNNs. In this section, two methods are introduced for multi-layer BNNs: \textbf{Hybrid-BNN} and \textbf{\texttt{DM}-BNN}, to take the advantages of \texttt{DM} strategy. A $4$-voter two-layer BNN (\textit{i.e.} with one hidden layer) is taken as an example, which has been well trained with the distribution parameters $(\sigma_{1}, \mu_{1})$ and $(\sigma_{2}, \mu_{2})$ for each layer.

The dataflows of these two methods are illustrated in Fig.~\ref{fig:deep-bnn}. For the Hybrid-BNN method, the \texttt{DM} strategy is only applied in the first layer and the corresponding dataflows are shown in Fig.~\ref{fig:deep-bnn}(a) which could be derived from Fig.~\ref{fig:df-naive} and Fig.~\ref{fig:df-dm}. In the first layer, $\beta_{11}$ and $\eta_{11}$ are pre-computed using $\bm{x}$, $\sigma_1$ and $\mu_1$, and memorized. Four sampled uncertainty matrices ($H_{11} \sim H_{14}$) are processed with $\beta_{11}$ and $\eta_{11}$ using \texttt{DM} strategy yielding four outputs ($\bm{y}_{11} \sim \bm{y}_{14}$). In the second layer, four weight matrices ($W_{21} \sim W_{24}$) are obtained based on scale-location transformation and operated with the previous four outputs ($\bm{y}_{11} \sim \bm{y}_{14}$), respectively. And then the results ($\bm{y}_{21} \sim \bm{y}_{24}$) are obtained for voting to determine the finally BNN response $\bar{\bm{y}}$.

For the \texttt{DM}-BNN method, the \texttt{DM} strategy is applied in all layers as graphically depicted in Fig.~\ref{fig:deep-bnn}(b). The $1$-input to $T$-outputs relationship now exists not only in the first layer but also the following layer(s). In the first layer, two uncertainty matrices ($H_{11}$ and $H_{12}$) are sampled and two corresponding outputs ($\bm{y}_{11}$ and $\bm{y}_{12}$) are calculated using the \texttt{DM} strategy. In the second layer, $\bm{y}_{11}$ is treated as input and two output ($\bm{y}_{21}$ and $\bm{y}_{22}$) are generated based on the \texttt{DM} strategy. Similarly, two outputs ($\bm{y}_{23}$ and $\bm{y}_{24}$) corresponding to $\bm{y}_{12}$ are generated.

\subsection{Discussions}  

The performance of single-layer BNN and multi-layer BNN with \texttt{DM} strategy, as well as the memory overhead introduced by \texttt{DM} strategy are analyzed in this section.

\subsubsection{Single-layer BNN} To get inside on the implications of our proposal we further evaluate the required computation complexity in terms of number of operations, \textit{i.e.}, additions (ADD) and multiplications (MUL), and memory requirements.
%The efficiency and disadvantages of the proposed dataflow are analyzed in this section.

\begin{table}[]
\centering
\caption{Single-layer BNN Computation Cost.
% between naive BNN ( Algorithm~\ref{alg:bnn})  and  Algorithm~\ref{alg:dfbnn}.
}
\label{table:computation}
\begin{tabular}{|c|c|c|}
\specialrule{1pt}{0pt}{0pt}
\multicolumn{3}{|c|}{Computation cost without \texttt{DM} strategy (Algorithm 1)}                                   \\
\hline
Operation                     & \# MUL & \# ADD            \\
\hline
$Q_k = H_k \times \sigma$     &   $MNT$ &  0                \\
$W_k = Q_k + \mu$             &   0     &  $MNT$            \\
$\bm{y_k} = W_k \cdot \bm{x}$ & $MNT$   &  $M(N-1)T$        \\
\hline
Total                         & $2MNT$  & $\approx 2MNT$    \\
\specialrule{0.8pt}{0pt}{0pt}
\specialrule{0.8pt}{0pt}{0pt}
\multicolumn{3}{|c|}{Computation cost with \texttt{DM} strategy (Algorithm 2)}                                       \\
\hline
Operation                       & \# MUL & \# ADD             \\
\hline
 $\eta = \mu \cdot \bm{x}$      &  $MN$     &   0             \\
 $\beta = \sigma \times \bm{x}$ &  $MN$     &  $M(N-1)$       \\
 $\bm{z_k} = <H_k,\beta>_L$     &  $MNT$    &  $M(N-1)T$      \\
 $\bm{y_k} = \bm{z_k} + \eta$   &  0        &  $MT$           \\
 \hline
 Total                          &  $MN(T+2)$&  $\approx MN(T+1)$     \\
 \specialrule{1pt}{0pt}{0pt}
\end{tabular}
\end{table}

Table~\ref{table:computation} summarizes the number of arithmetic operation required by the  dataflow in Fig.~\ref{fig:df-naive} and Fig.~\ref{fig:df-dm}, \textit{i.e.}, Algorithm~\ref{alg:bnn} and Algorithm~\ref{alg:dfbnn}, respectively. If we concentrate on the number of multiplications, as they are more time consuming, the two approaches require $2MNT$ and $MN(T+2)$ multiplications, respectively. This indicates that if $T > 2$, which is obviously the case if one would like to deal with uncertainties by means of a Bayesian approach, the \texttt{DM} approach outperforms standard one. And as $T$ increases the advantage is more substantial, reaching a theoretical maximum of 50\% reduction as suggested by Eqn.~\eqref{Equation:improvement}:
\begin{equation}
\label{Equation:improvement}
\lim_{T \to \infty} \frac{MN(T+2)}{2MNT} = \frac{1}{2}.
\end{equation}

Given that one addition takes one cycle and one multiplication by $2$ cycles in state of the art processors, the computation cost in terms of ADD only could be regarded as $\approx 3MNT$ and $\approx 6MNT$ for single-layer BNN evaluation with and without \texttt{DM} strategy, respectively, which results in an overall speedup of $\approx 2$. 

\subsubsection{Multi-layer BNN} 
Related to the multi-layer BNN dataflows in Fig.~\ref{fig:deep-bnn}, the performance of Hybrid-BNN and \texttt{DM}-BNN will be discussed.

In Hybrid-BNN dataflow, the \texttt{DM} strategy is only applied in the first layer. Thus, it could only reduce the computation cost of the first layer. Usually, the first layer accounts for more than 80\% of the total computation, and consequently the computation cost of the entire flow in Hybrid-BNN could be still reduced by 40\%. In \texttt{DM}-BNN dataflow, the computation cost of all layers could be reduce by about 50\%.  It can also be found that the required number of uncertainty matrices for each layer in \texttt{DM}-BNN is less than that in Hybrid-BNN. As illustrated in Fig.~\ref{fig:deep-bnn}, 8 uncertainty matrices (4 for each layer) are sampled in Hybrid-BNN in order to get 4 voter results. While 4 uncertainty matrices (2 for each layer) are sampled in \texttt{DM}-BNN. In general, if BNN has $\mathcal{L}$ layers, $\sqrt[\mathcal{L}]{T}$ uncertainty matrices are sufficient for each layer to obtain $T$ voter results. Less uncertainty matrices mean less computation cost. Finally, \texttt{DM}-BNN could reduce more than 50\% computation cost when compared with standard BNN. Even though some voting outputs inherit the same uncertainty, the experimental results indicate that its influence could be ignored.

\subsubsection{\texttt{DM} deployment  in convolutional layers} \label{Section:bnn:dis:cnn}

Convolutional Neural Networks (CNNs) are a class of deep neural networks able to capture spatial and temporal dependencies in an image through the application of relevant filters. The advantages of CNNs make them most successful in perspective tasks. Thus, it is important to extend the proposed \texttt{DM} strategy in convolutional layers. Fortunately, this extension could be achieved by means of convolutional layer unfolding~\cite{chellapilla2006high} which is a well known technique that has been utilized in many applications (e.g.~\cite{chetlur2014cudnn,lecun2010convolutional,ciresan2011flexible}). This technique relies on the creation of convolution matrices such that the the convolution computation is transformed into a matrix multiplication. Thus, after applying unfolding on the convolution layers the DM strategy can be directly applied to them.

% which is commonly implemented in signal processing and communications applications.

\subsubsection{Memory overhead} 
Based on the previous analysis the proposed \texttt{DM} strategy could effectively reduce the computation cost at the expense of  additional local memory. In the standard BNN dataflow, we only need to store the weight distribution parameters $\sigma$ and $\mu$. But for \texttt{DM} dataflow, additional storage is required for matrix $\beta$ who has the same dimension as $\sigma$ and $\mu$, which means about 50\% memory overhead is introduced.
 \iffalse

 \bibliography{../ref/BNN}

 \fi

\section{Memory reduction for \texttt{DM} strategy} \label{Section:memory}

\begin{figure}[tb!]
    \centering
    \includegraphics[width=0.45\textwidth]{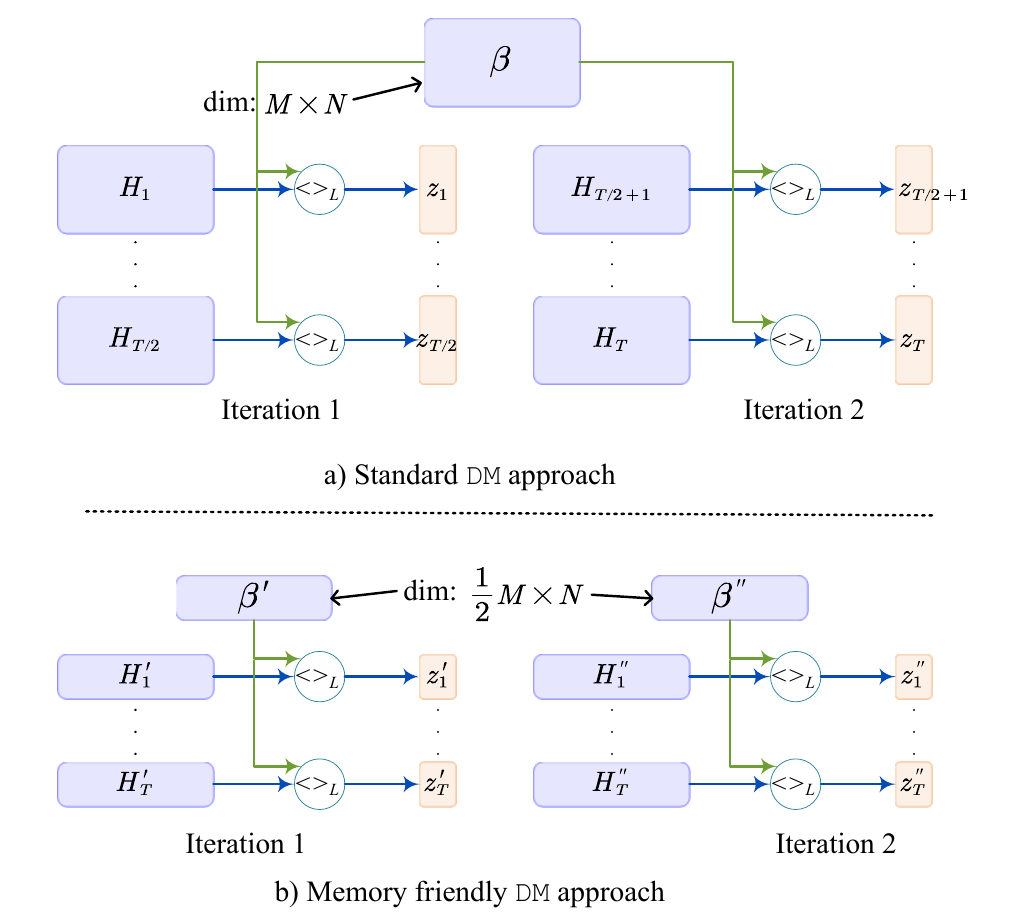}
    \caption{ BNN evaluation with $\alpha=1/2$ for memory reduction. a) Standard \texttt{DM} approach. b) Memory friendly \texttt{DM} approach.  The additional memory allocated by b) is reduced from $M \times N$ to $\alpha M \times N$ (i.e., by $\frac{1}{2}M \times N$). In b), variable with $'$ indicates the first half part of it and with $''$ indicates the second half of it. 
    For example, $\beta  = \left[ {\beta '} {\beta '}\right]^{T} $. 
    }
    \label{fig:mem-opt}
\end{figure}
In this section, the memory overhead issue introduced by \texttt{DM} strategy is considered and a memory friendly computing mechanism is proposed. The goal is to minimize the additional allocated memory as much as possible without inducing  additional computation overhead. 

Generally speaking, due to hardware limitations one cannot simultaneously perform the evaluation of all the $T$-sampling feed-forward neural networks. For some edge devices, a neural network is even divided into several parts and operates in a time-multiplexed manner~\cite{ren2016designing,indiveri2015neuromorphic}. When assuming that only $\alpha T$-sampling feed-forward neural networks can be evaluated simultaneously where $0 < \alpha \le 1$, only $\alpha T M N$ Gaussian random numbers are sampled in each iteration. Thus, $\alpha T$ entire uncertainty matrices ($H \in \mathbb{R}^{M \times N}$) and voting outputs ($\bm{y} \in \mathbb{R}^{M}$) are generated in each iteration while ${\alpha}^{-1}$ iterations are required totally. In order to match the dimension of uncertainty matrix, additional memory with size of $M \times N$ should be allocated to store $\beta$ by this way. 

If we want to reduce the memory overhead of \texttt{DM} strategy, the dimension of uncertainty matrix involved in each iteration should be shrunken first. Guided by this, a memory friendly computing mechanism can be derived. In this computing mechanism, $\alpha T M N$ Gaussian random numbers generated in each iteration are redistributed to $T$ sub-matrices ($H' \in \mathbb{R}^{\alpha M \times N}$). And consequently $\beta$ could be partially computed and memorized with the same size as $H'$. At the end of each iteration, $T$ sub-outputs ($\bm{y}' \in \mathbb{R}^{\alpha M}$) are calculated, and  after ${\alpha}^{-1}$ iterations, all $T$ voter outputs are obtained. With this approach, the introduced memory overhead by \texttt{DM} strategy could be reduced from $50\%$ to $\alpha \times 50\%$. For the sake of concise and simplicity, the mechanism is illustrated for $\alpha=1/2$ in Fig.~\ref{fig:mem-opt}. 

% Version 1
% It is worth to note that the benefit of proposed memory-friendly \texttt{DM} approach is determined by $\alpha$. If $\alpha = 1$ which means that all $T$ voters are evaluated in parallel, there will be no advantage for memory-friendly \texttt{DM} approach. But in practical implementations it will seldom happen because hardware resources are usually limited. 

% Version 2
It is worth to note that the benefit of the proposed memory friendly approach is determined by computation resources. 
If only limited computation resources are available, it could reduce the memory overhead as much as possible while maintaining the computation performance. If the computation resources is adequate that all $T$ voters could be evaluated in parallel, the memory friendly approach could provide a trade-off between computation performance and memory overhead.

 \iffalse

 \bibliography{../ref/BNN}

 \fi

\section {Experimental Results} \label{Section:experiment}

\begin{figure}[tb!]
    \centering
    \includegraphics[width=0.5\textwidth]{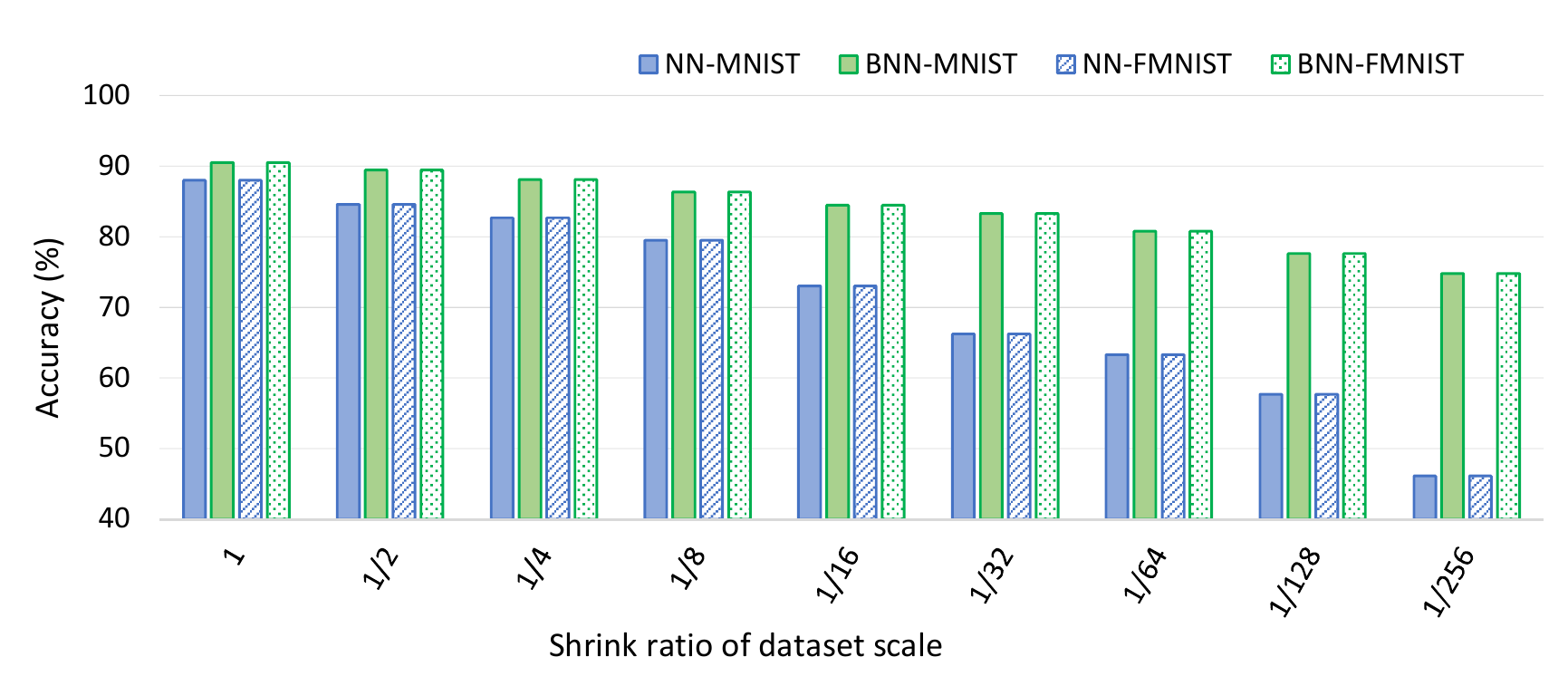}
    \caption{Accuracy comparison between NN and BNN on MNIST/FMNIST datesets for different training data scale. The accuracy is calculated using 10000 testing datasets.}
    \label{fig:small-data-mnist}
\end{figure}

% \begin{figure}[tb!]
%     \centering
%     \includegraphics[width=0.45\textwidth]{small-data-fmnist.pdf}
%     \caption{{\color{blue}Training accuracy of FMNIST dateset between NN and BNN with different training data scale.}}
%     \label{fig:small-data-fmnist}
% \end{figure}
% }

The advantages of BNN have been demonstrated by many previous works. The work of~\cite{kendall2017uncertainties,zhu2018bayesian,van2017bayesian} show BNNs ability when dealing with uncertainty. The advantages of BNNs in autonomous vehicle safety is presented in~\cite{mcallister2017concrete,feng2018towards}. Work~\cite{cai2018vibnn} suggests that BNN performs much better than DNN as training data size shrinks. Rawat et al.~\cite{rawat2017adversarial} conclude that BNN can be considered for detecting adversarial examples. In this section, we first demonstrate that the strengths of BNN in classification tasks using small datasets. Then the efficiency of the proposed \texttt{DM} strategy are evaluated.

\subsection {BNN Performance on small-scale datasets} \label{Section:experiment:small}

In order to evaluate BNN performance on small datasets, two famous datasets MNIST~\cite{mnist} and FMNIST~\cite{fmnist} are utilized. Both MNIST and FMNIST datasets have 60000 images for training and 10000 images for testing. Each image is a gray bitmap whose size is 28 $\times$ 28. The  MNIST dataset contains 10 classes of handwritten digits (from `0' to `9'), while FMNIST dataset contains 10 classes of clothes or shoes. In this experiment, two datesets are reduced to samll-scale datasets based on a shrink ratio. In the reduced datasets, the images are randomly selected from the original dataset and the number of images for each class are the same. For example with the shrink ratio of 256, each class has about $24$ (\textit{i.e.}, $\lceil 60000/256/10 \rceil$) images to be chosen as a subset. Due to that there are 10 classes in MNIST/FMNIST, each subset contains 240 images for training. It is noted that there are still 10000 images for testing no matter how many training images there are. 

For the MNIST dateset, a $3$-layer (with 2 hidden layers) fully connected neural network is built with a $784$-$200$-$200$-$10$ configuration. For FMNIST dateset which is much more complicated, the LeNet-5~\cite{lenet5} structure is utilized. The non-Bayesian neural network is trained using TensorFlow \cite{abadi2016tensorflow} while the Bayesian neural network is trained using Edward framework \cite{tran2017edward}. The training parameters such as epochs, batch size and learning rate are set to be the same for fairness. An overview for the NN and BNN achieved accuracy on the small-scale datasets is presented in Fig.~\ref{fig:small-data-mnist}. The horizontal axis represents the shrink ratio of original dataset and the vertical axis represents the testing accuracy on the 10000 testing datasets. The figure clearly indicates that BNN could achieve better performance than non-Bayesian neural network when the training date size shrinks.

\subsection {\texttt{DM} strategy evaluation}

In this work, two methods are described for multi-layer BNNs in Section~\ref{Section:computation}, which are theoretically proved to reduce the computation complexity. In this section, both software implementation and hardware implementation are presented in order to evaluate the performance of the proposed \texttt{DM} strategy.
The $3$-layer (with 2 hidden layers) fully connected neural network with MNIST dataset are used. The number of sampling $T$ in standard BNN and Hybrid-BNN for each layer is set as $100$. And for the three layers in \texttt{DM}-BNN, $T$ is set as $10$, $10$, and $5$, respectively. And consequently $500$ voting results are generated for \texttt{DM}-BNN.

In this experiments, both (1) Hybrid-BNN: the \texttt{DM} strategy is applied only in the first layer, and (2) \texttt{DM}-BNN: the \texttt{DM} strategy is applied in all layers, are implemented for evaluation. Recently, Cai et al.~\cite{cai2018vibnn} propose an approach to accelerate BNN inference. In their work, two GRNGs and some architecture level optimizations are proposed. Its dataflow is also re-implemented in our experiments which is referred as standard BNN (the \texttt{DM} strategy is not applied at all). In order to make a fair comparison, the energy consumption of GRNGs is not calculated and no architecture level optimizations are implemented. By this way, the comparison results could well demonstrate the efficiency of the proposed \texttt{DM} strategy.

\subsubsection {Software Implementation}

All methods are implemented using Python language and evaluated on a $64$-bit Linux server. Table~\ref{table:software} summarizes the accuracy and required number of operations. The second column reports the prediction accuracy for the $10000$ MNIST testing images, while the computation complexity in terms of multiplications (\# MUL) and additions (\# ADD) is given in column 3 and column 4, respectively. 

\begin{table}[tb]
\centering
\caption{Software Implementation Results.}
\label{table:software}
\begin{tabular}{c|c|c|c}
    \specialrule{0.8pt}{0pt}{0pt}
     Method & Accuracy & \# MUL ($\times 10^6$) & \# ADD ($\times 10^6$) \\
    \hline
     Standard BNN    & 96.73\%  & 39.8  & 39.7 \\
     Hybrid-BNN    & 96.73\%   & 24.2   & 24.1    \\
     \texttt{DM}-BNN    &   96.7\%      & 6.9  & 6.7  \\
     \specialrule{0.8pt}{0pt}{0pt}
\end{tabular}
\end{table}

In Hybrid-BNN, the \texttt{DM}  strategy is only applied in the first layer, which covers about 79\% of total network computation,  and the Table~\ref{table:software} confirms the expected Hybrid-BNN theoretical computation reduction of about 39\%. 
\texttt{DM}-BNN could effectively reduce the computation cost because feature decomposition and memorization strategy is utilized with all layers. Moreover, because only $10$ uncertainty matrices are sampled to participate in the first BNN layer, the total computation dropped by 82.5\%, which is beyond the 50\% upper bound achievable by \texttt{DM}  in single-layer BNNs. 
From Table~\ref{table:software} we can observe that the accuracy of Hybrid-BNN and standard BNN are the same as expected, while \texttt{DM}-BNN could significantly reduce the computations only at the cost of very slight accuracy degradation.

\subsubsection {Hardware Implementation} \label{Section:experiment:dm:hw}
Hardware implementations of the three methods are also performed for evaluating the BNN inference latency, power consumption and area efficiency. They are implemented as Verilog and synthesized by Synopsys Design Compiler with $45$ $nm$ FreePDK technology. $8$-bit fixed point number representation is utilized in the designs. Cacti~\cite{muralimanohar2007CACTI} is exploited to estimate the area and energy consumption of the involved memory.

\begin{figure}[tb!]
    \centering
    \includegraphics[width=0.4\textwidth]{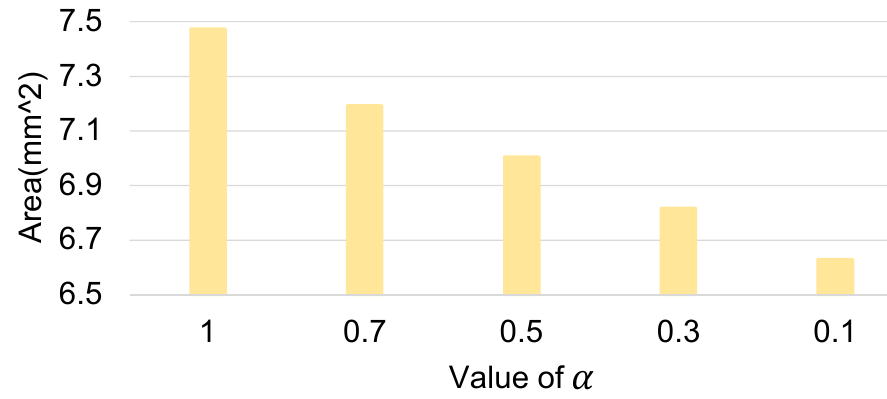}
    \caption{Evaluation of memory reduction strategy.}
    \label{fig:mem-opt-area}
\end{figure}

We first evaluate the efficiency of the memory-friendly computing mechanism, which is proposed to reduce the memory overhead introduced by \texttt{DM} strategy. 
This mechanism is designed based on the fact that hardware resources are always limited.
So in this experiments, the computing mechanism efficiency is evaluated by means of required system area. 
Assuming that only $\alpha T$-sampling feed-forward neural networks can be evaluated simultaneously, Fig.~\ref{fig:mem-opt-area} presents the system area corresponding to different $\alpha$ values. The horizontal axis means the value of $\alpha$.  And the vertical axis means the required hardware system area ($mm^2$). The experimental results demonstrate that when $\alpha$ decreases, the required hardware system area could also be reduced.

The efficiency of the proposed \texttt{DM} strategy in BNN inference stage is evaluated with $\alpha = 0.1$.
Table~\ref{table:hardware} describes the comparison of the three hardware implementations in terms of accuracy, area ($mm^2$), energy consumption ($nJ$), and total execution time ($\times 10^3 s$) for BNN inference. One can observe that the accuracy has been slightly diminished when compared with the software implementation but this is inherent due to the utilization of lower precision,  $8$-bit fixed point number instead of 32-bit floating point number. 
Compared with the standard BNN, Hybrid-BNN and \texttt{DM}-BNN has about 27\% and 14\% area overhead, respectively. There are two reasons why standard BNN has a best area efficiency while Hybrid-BNN has a worst area efficiency. First, both Hybrid-BNN and \texttt{DM}-BNN require extra local memory which is inherent to  feature decomposition and memorization strategy utilization. Second, while  hardware resources could be shared among different layers in standard BNN and \texttt{DM}-BNN, because the computing mechanisms of all layers are the same, this is not the case for Hybrid-BNN, which first layer requires a different mechanism than the other layers. As reported in column 4, Hybrid-BNN and \texttt{DM}-BNN provide 29\% and 73\% energy consumption reductions, respectively when compared with standard BNN. Concerning the total execution time required for the evaluation of one MNIST test data, a speedup of  1.5$\times$ and 4$\times$ is obtained for Hybrid-BNN and \texttt{DM}-BNN over the standard BNN, respectively.

To summarize, Hybrid-BNN reduces the energy consumption by 29\% and achieves a 1.5$\times$ speedup at the expense 27\% overhead in area,  while \texttt{DM}-BNN reduces the energy consumption by 73\% and achieves a 4$\times$ speedup at the expense 14\% overhead with a slight accuracy decreasing.

\begin{table}[tb]
\centering
\caption{Hardware Implementation Results.}
\label{table:hardware}
\begin{tabular}{c|c|c|c|c}
    \specialrule{0.8pt}{0pt}{0pt}
     Method & Accuracy & \tabincell{c}{Area \\ ($mm^2$)} & \tabincell{c}{Energy \\ ($\mu J$)}  & \tabincell{c}{Runtime \\ ($\mu s$)} \\
    \hline
     Standard BNN    & 95.42\%    & 5.76   & 172    & 392 \\
     Hybrid-BNN    & 95.42\%     &  7.33   & 122     & 259  \\
     \texttt{DM}-BNN    &  95.35\%    & 6.63     & 46   & 97 \\
     \specialrule{0.8pt}{0pt}{0pt}
\end{tabular}
\end{table}
 \iffalse

 \bibliography{../ref/Bayesian}

\begin{thebibliography}{10}
\providecommand{\url}[1]{#1}
\csname url@samestyle\endcsname
\providecommand{\newblock}{\relax}
\providecommand{\bibinfo}[2]{#2}
\providecommand{\BIBentrySTDinterwordspacing}{\spaceskip=0pt\relax}
\providecommand{\BIBentryALTinterwordstretchfactor}{4}
\providecommand{\BIBentryALTinterwordspacing}{\spaceskip=\fontdimen2\font plus
\BIBentryALTinterwordstretchfactor\fontdimen3\font minus
  \fontdimen4\font\relax}
\providecommand{\BIBforeignlanguage}[2]{{%
\expandafter\ifx\csname l@#1\endcsname\relax
\typeout{** WARNING: IEEEtran.bst: No hyphenation pattern has been}%
\typeout{** loaded for the language `#1'. Using the pattern for}%
\typeout{** the default language instead.}%
\else
\language=\csname l@#1\endcsname
\fi
#2}}
\providecommand{\BIBdecl}{\relax}
\BIBdecl

\bibitem{lecun1995convolutional}
Y.~LeCun, Y.~Bengio \emph{et~al.}, ``Convolutional networks for images, speech,
  and time series,'' \emph{The Handbook of Brain Theory and Neural Networks},
  vol. 3361, no.~10, p. 1995, 1995.

\bibitem{hochreiter1997long}
S.~Hochreiter and J.~Schmidhuber, ``Long short-term memory,'' \emph{Neural
  computation}, vol.~9, no.~8, pp. 1735--1780, 1997.

\bibitem{goodfellow2014generative}
I.~Goodfellow, J.~Pouget-Abadie, M.~Mirza, B.~Xu, D.~Warde-Farley, S.~Ozair,
  A.~Courville, and Y.~Bengio, ``Generative adversarial nets,'' in
  \emph{Proc.~NIPS}, 2014, pp. 2672--2680.

\bibitem{krizhevsky2012imagenet}
A.~Krizhevsky, I.~Sutskever, and G.~E. Hinton, ``{ImageNet} classification with
  deep convolutional neural networks,'' in \emph{Proc.~NIPS}, 2012, pp.
  1097--1105.

\bibitem{sutskever2014sequence}
I.~Sutskever, O.~Vinyals, and Q.~V. Le, ``Sequence to sequence learning with
  neural networks,'' in \emph{Proc.~NIPS}, 2014, pp. 3104--3112.

\bibitem{hinton2012deep}
G.~Hinton, L.~Deng, D.~Yu, G.~Dahl, A.-r. Mohamed, N.~Jaitly, A.~Senior,
  V.~Vanhoucke, P.~Nguyen, B.~Kingsbury \emph{et~al.}, ``Deep neural networks
  for acoustic modeling in speech recognition,'' \emph{IEEE Signal Processing
  Magazine}, vol.~29, 2012.

\bibitem{chen2015deepdriving}
C.~Chen, A.~Seff, A.~Kornhauser, and J.~Xiao, ``Deepdriving: Learning
  affordance for direct perception in autonomous driving,'' in
  \emph{Proc.~ICCV}, 2015, pp. 2722--2730.

\bibitem{he2016deep}
K.~He, X.~Zhang, S.~Ren, and J.~Sun, ``Deep residual learning for image
  recognition,'' in \emph{Proc.~CVPR}, 2016, pp. 770--778.

\bibitem{bishop2006pattern}
C.~M. Bishop, \emph{Pattern recognition and machine learning}.\hskip 1em plus
  0.5em minus 0.4em\relax Springer, 2006.

\bibitem{nguyen2015deep}
A.~Nguyen, J.~Yosinski, and J.~Clune, ``Deep neural networks are easily fooled:
  High confidence predictions for unrecognizable images,'' in
  \emph{Proc.~CVPR}, 2015, pp. 427--436.

\bibitem{gal2015bayesian}
Y.~Gal and Z.~Ghahramani, ``Bayesian convolutional neural networks with
  bernoulli approximate variational inference,'' \emph{arXiv preprint
  arXiv:1506.02158}, 2015.

\bibitem{ticknor2013bayesian}
J.~L. Ticknor, ``A {Bayesian} regularized artificial neural network for stock
  market forecasting,'' \emph{Journal of Expert Systems with Applications},
  vol.~40, no.~14, pp. 5501--5506, 2013.

\bibitem{spinbis}
X.~Jia, J.~Yang, P.~Dai, R.~Liu, Y.~Chen, and W.~Zhao, ``{SPINBIS}:
  Spintronics-based {B}ayesian inference system with stochastic computing,''
  \emph{IEEE TCAD}, vol.~39, no.~4, pp. 789--802, 2020.

\bibitem{jia2017spintronics}
X.~Jia, J.~Yang, Z.~Wang, Y.~Chen, and W.~Zhao, ``Spintronics based stochastic
  computing for efficient {Bayesian} inference system,'' in
  \emph{Proc.~ASPDAC}, 2018, pp. 580--585.

\bibitem{chien2016bayesian}
J.-T. Chien and Y.-C. Ku, ``Bayesian recurrent neural network for language
  modeling,'' \emph{IEEE TNNLS}, vol.~27, no.~2, pp. 361--374, 2016.

\bibitem{tran2017edward}
D.~Tran, M.~D. Hoffman, R.~A. Saurous, E.~Brevdo, K.~Murphy, and D.~M. Blei,
  ``Deep probabilistic programming,'' in \emph{Proc.~ICLR}, 2017.

\bibitem{bingham2018pyro}
E.~Bingham, J.~P. Chen, M.~Jankowiak, F.~Obermeyer, N.~Pradhan, T.~Karaletsos,
  R.~Singh, P.~Szerlip, P.~Horsfall, and N.~D. Goodman, ``{Pyro}: Deep
  universal probabilistic programming,'' \emph{arXiv preprint
  arXiv:1810.09538}, 2018.

\bibitem{shi2017zhusuan}
J.~Shi, J.~Chen, J.~Zhu, S.~Sun, Y.~Luo, Y.~Gu, and Y.~Zhou, ``{ZhuSuan}: A
  library for {Bayesian} deep learning,'' \emph{arXiv preprint
  arXiv:1709.05870}, 2017.

\bibitem{han2015learning}
S.~Han, J.~Pool, J.~Tran, and W.~Dally, ``Learning both weights and connections
  for efficient neural network,'' in \emph{Proc.~NIPS}, 2015, pp. 1135--1143.

\bibitem{denton2014exploiting}
E.~L. Denton, W.~Zaremba, J.~Bruna, Y.~LeCun, and R.~Fergus, ``Exploiting
  linear structure within convolutional networks for efficient evaluation,'' in
  \emph{Proc.~NIPS}, 2014, pp. 1269--1277.

\bibitem{wen2016learning}
W.~Wen, C.~Wu, Y.~Wang, Y.~Chen, and H.~Li, ``Learning structured sparsity in
  deep neural networks,'' in \emph{Proc.~NIPS}, 2016, pp. 2074--2082.

\bibitem{knag2015sparse}
P.~Knag, J.~K. Kim, T.~Chen, and Z.~Zhang, ``A sparse coding neural network
  {ASIC} with on-chip learning for feature extraction and encoding,''
  \emph{IEEE JSSC}, vol.~50, no.~4, pp. 1070--1079, 2015.

\bibitem{cai2018vibnn}
R.~Cai, A.~Ren, N.~Liu, C.~Ding, L.~Wang, X.~Qian, M.~Pedram, and Y.~Wang,
  ``{VIBNN}: {Hardware} acceleration of {Bayesian} neural networks,'' in
  \emph{Proc.~ASPLOS}.\hskip 1em plus 0.5em minus 0.4em\relax ACM, 2018, pp.
  476--488.

\bibitem{graves2011practical}
A.~Graves, ``Practical variational inference for neural networks,'' in
  \emph{Proc.~NIPS}, 2011, pp. 2348--2356.

\bibitem{blundell2015weight}
C.~Blundell, J.~Cornebise, K.~Kavukcuoglu, and D.~Wierstra, ``Weight
  uncertainty in neural networks,'' \emph{arXiv preprint arXiv:1505.05424},
  2015.

\bibitem{chen2014stochastic}
T.~Chen, E.~Fox, and C.~Guestrin, ``Stochastic gradient {Hamiltonian Monte
  Carlo},'' in \emph{Proc.~ICML}, 2014, pp. 1683--1691.

\bibitem{balan2015bayesian}
A.~K. Balan, V.~Rathod, K.~P. Murphy, and M.~Welling, ``Bayesian dark
  knowledge,'' in \emph{Proc.~NIPS}, 2015, pp. 3438--3446.

\bibitem{malik2016gaussian}
J.~S. Malik and A.~Hemani, ``Gaussian random number generation: A survey on
  hardware architectures,'' \emph{ACM Computing Surveys (CSUR)}, vol.~49,
  no.~3, p.~53, 2016.

\bibitem{thomas2007gaussian}
D.~B. Thomas, W.~Luk, P.~H. Leong, and J.~D. Villasenor, ``Gaussian random
  number generators,'' \emph{ACM Computing Surveys (CSUR)}, vol.~39, no.~4,
  p.~11, 2007.

\bibitem{chellapilla2006high}
K.~Chellapilla, S.~Puri, and P.~Simard, ``High performance convolutional neural
  networks for document processing,'' in \emph{Tenth International Workshop on
  Frontiers in Handwriting Recognition}, 2006.

\bibitem{chetlur2014cudnn}
S.~Chetlur, C.~Woolley, P.~Vandermersch, J.~Cohen, J.~Tran, B.~Catanzaro, and
  E.~Shelhamer, ``cudnn: Efficient primitives for deep learning,'' \emph{arXiv
  preprint arXiv:1410.0759}, 2014.

\bibitem{lecun2010convolutional}
Y.~LeCun, K.~Kavukcuoglu, and C.~Farabet, ``Convolutional networks and
  applications in vision,'' in \emph{Proc.~ISCAS}, 2010, pp. 253--256.

\bibitem{ciresan2011flexible}
D.~C. Ciresan, U.~Meier, J.~Masci, L.~M. Gambardella, and J.~Schmidhuber,
  ``Flexible, high performance convolutional neural networks for image
  classification,'' in \emph{Proc.~IJCAI}, 2011.

\bibitem{ren2016designing}
A.~Ren, Z.~Li, Y.~Wang, Q.~Qiu, and B.~Yuan, ``Designing reconfigurable
  large-scale deep learning systems using stochastic computing,'' in
  \emph{Proc.~ICRC}, 2016, pp. 1--7.

\bibitem{indiveri2015neuromorphic}
G.~Indiveri, F.~Corradi, and N.~Qiao, ``Neuromorphic architectures for spiking
  deep neural networks,'' in \emph{Proc.~IEDM}, 2015, pp. 4--2.

\bibitem{kendall2017uncertainties}
A.~Kendall and Y.~Gal, ``What uncertainties do we need in bayesian deep
  learning for computer vision?'' in \emph{Proc.~NIPS}, 2017, pp. 5574--5584.

\bibitem{zhu2018bayesian}
Y.~Zhu and N.~Zabaras, ``Bayesian deep convolutional encoder--decoder networks
  for surrogate modeling and uncertainty quantification,'' \emph{Journal of
  Computational Physics}, vol. 366, pp. 415--447, 2018.

\bibitem{van2017bayesian}
J.~van~der Westhuizen and J.~Lasenby, ``Bayesian {LSTMs} in medicine,''
  \emph{arXiv preprint arXiv:1706.01242}, 2017.

\bibitem{mcallister2017concrete}
R.~McAllister, Y.~Gal, A.~Kendall, M.~Van Der~Wilk, A.~Shah, R.~Cipolla, and
  A.~V. Weller, ``Concrete problems for autonomous vehicle safety: advantages
  of bayesian deep learning,'' in \emph{Proc.~IJCAI}, 2017, pp. 4745--4753.

\bibitem{feng2018towards}
D.~Feng, L.~Rosenbaum, and K.~Dietmayer, ``Towards safe autonomous driving:
  Capture uncertainty in the deep neural network for lidar {3D} vehicle
  detection,'' in \emph{Proc.~ITSC}, 2018, pp. 3266--3273.

\bibitem{rawat2017adversarial}
A.~Rawat, M.~Wistuba, and M.-I. Nicolae, ``Adversarial phenomenon in the eyes
  of {Bayesian} deep learning,'' \emph{arXiv preprint arXiv:1711.08244}, 2017.

\bibitem{mnist}
Y.~LeCun, C.~Cortes, and C.~J. Burges. (2010) {MNIST} handwritten digit
  database. \url{http://yann.lecun.com/exdb/mnist}.

\bibitem{fmnist}
H.~Xiao, K.~Rasul, and R.~Vollgraf. (2017) Fashion-{MNIST}: a novel image
  dataset for benchmarking machine learning algorithms.

\bibitem{lenet5}
Y.~LeCun, L.~Bottou, Y.~Bengio, P.~Haffner \emph{et~al.}, ``Gradient-based
  learning applied to document recognition,'' \emph{Proceedings of the IEEE},
  vol.~86, no.~11, pp. 2278--2324, 1998.

\bibitem{abadi2016tensorflow}
M.~Abadi, A.~Agarwal, P.~Barham, E.~Brevdo, Z.~Chen, C.~Citro, G.~S. Corrado,
  A.~Davis, J.~Dean, M.~Devin \emph{et~al.}, ``Tensorflow: Large-scale machine
  learning on heterogeneous distributed systems,'' \emph{arXiv preprint
  arXiv:1603.04467}, 2016.

\bibitem{muralimanohar2007CACTI}
N.~Muralimanohar, R.~Balasubramonian, and N.~Jouppi, ``Optimizing {NUCA}
  organizations and wiring alternatives for large caches with {CACTI} 6.0,'' in
  \emph{Proc.~MICRO}, 2007, pp. 3--14.

\end{thebibliography}

 \fi

\section {Conclusions} \label{Section:conclusion}

This paper addresses the high computation complexity of BNN inference procedure and introduces a novel computation efficient BNN inference approach, which potentially enables BNNs' utilization in resources-constrained systems, e.g., IoT. We conduct a deep analysis of BNN inference dataflow and introduce a \texttt{DM} strategy that reduces the computation complexity, while having negligible BNN inference accuracy reduction, at the expense of some memory overhead. We further propose \texttt{DM} strategy utilization in multi-layer BNNs and introduce a memory friendly computing framework which is able to mitigate the \texttt{DM} induced memory overhead. Finally, our approach is implemented by Verilog and synthesized with $45$ $nm$ FreePDK technology. Evaluation results demonstrate that the proposed strategy provides an energy consumption reduction of 73\% and a 4$\times$ speedup at the expense of 14\% area overhead compared with the standard BNN inference.
As a final remark we note that the reported performance improvement can be further improved by means of architecture level optimization (e.g., memory optimization in~\cite{cai2018vibnn}) or network compression (e.g., pruning in~\cite{han2015learning}), which constitute future work subjects.

{
\bibliographystyle{IEEEtran}
% \bibliography{./Top_sim,./BNN}
% Generated by IEEEtran.bst, version: 1.14 (2015/08/26)

}

\vspace{-15mm}

\begin{IEEEbiography}[{\includegraphics[width=1in,height=1.25in,clip,keepaspectratio]{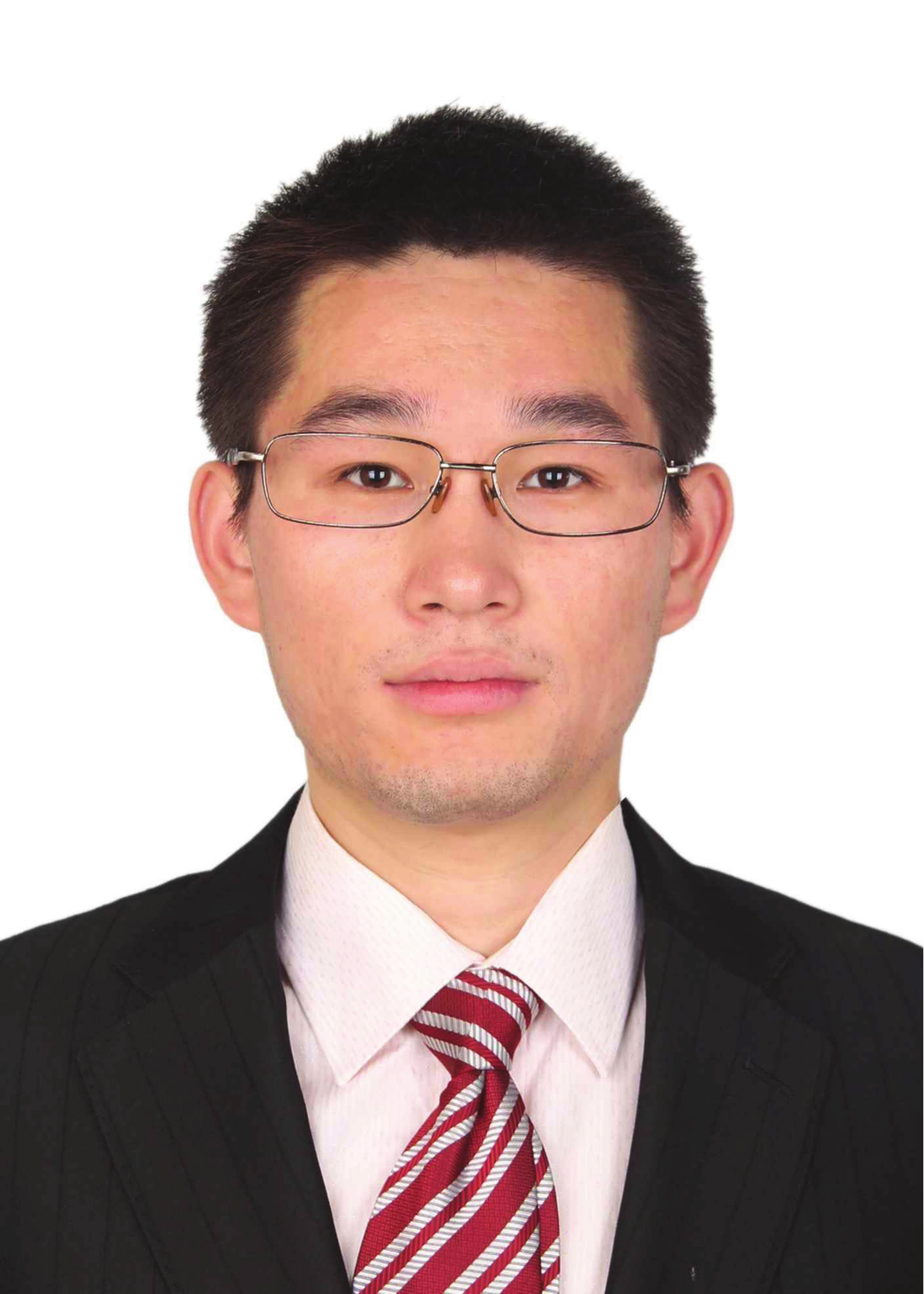}}]{Xiaotao Jia}

(S'13-M'17) received the B.S. degree in mathematics from Beijing Jiao Tong University, Beijing, China, in 2011, and the Ph.D. degree in computer science and technology from Tsinghua University, Beijing, China, in 2016.

He is currently an Assistant Professor with the School of Microelectronics in Beihang University, Beijing, China. From 2016 to 2019, he was a post-doctoral researcher with the School of Microelectronics in Beihang University. His current research interests include spintronic circuits, stochastic computing and Bayesian deep learning.

\end{IEEEbiography}

\vspace{-15mm}

\begin{IEEEbiography}[{\includegraphics[width=1in,height=1.25in,clip,keepaspectratio]{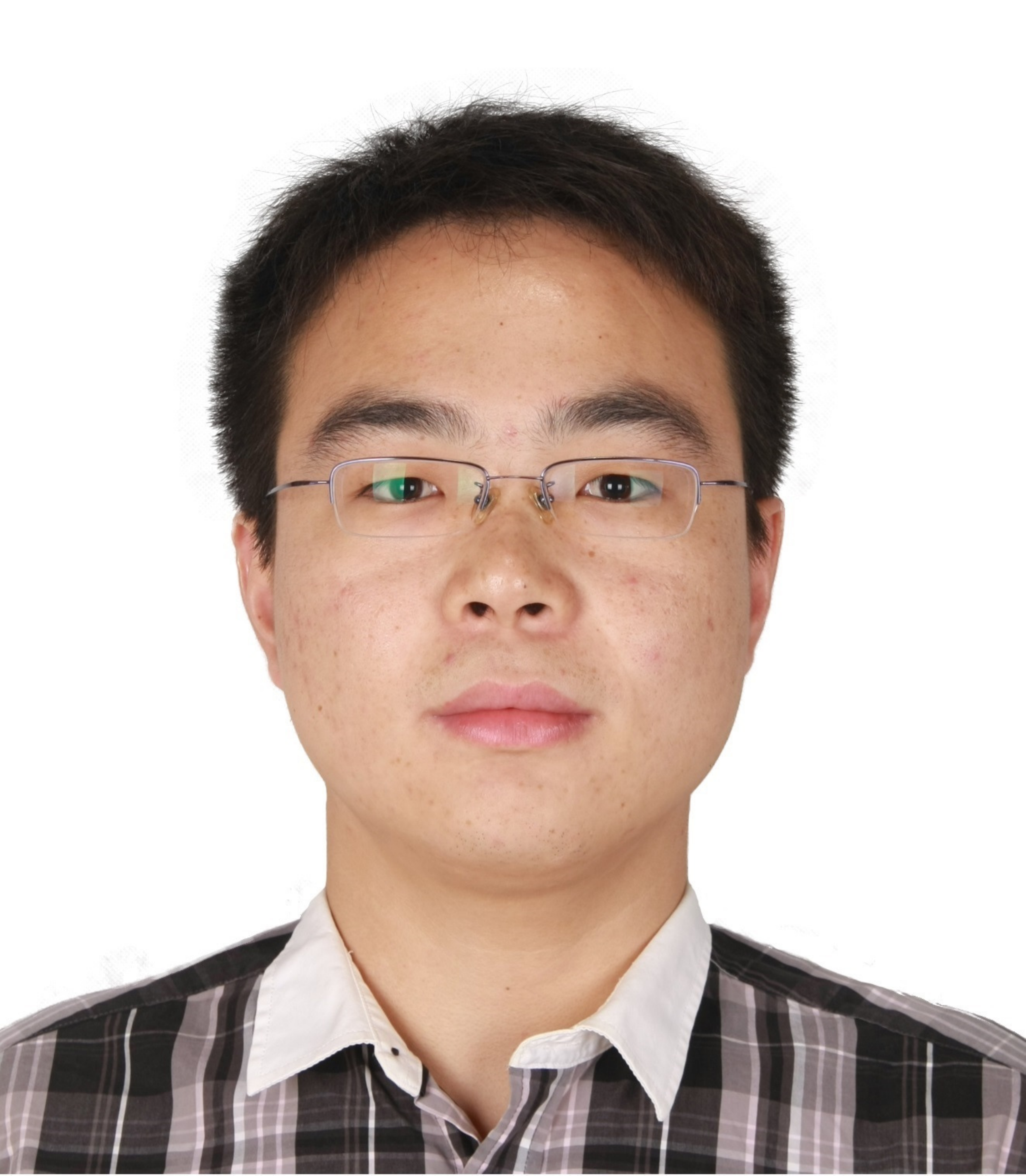}}]{Jianlei Yang}

(S'11-M'14) received the B.S. degree in microelectronics from Xidian University, Xi'an, China, in 2009, and the Ph.D. degree in computer science and technology from Tsinghua University, Beijing, China, in 2014.

He is currently an Associate Professor in Beihang University, Beijing, China, with the School of Computer Science and Engineering. From 2014 to 2016, he was a post-doctoral researcher with the Department of ECE, University of Pittsburgh, Pennsylvania, United States.
His current research interests include spintronics and neuromorphic computing systems.

Dr. Yang was the recipient of the First/Second place on ACM TAU Power Grid Simulation Contest in 2011/2012. He was a recipient of IEEE ICCD Best Paper Award in 2013, IEEE ICESS Best Paper Award in 2017, and ACM GLSVLSI Best Paper Nomination in 2015.

\end{IEEEbiography}

% \vspace{-15mm}

\begin{IEEEbiography}[{\includegraphics[width=1in,height=1.25in,clip,keepaspectratio]{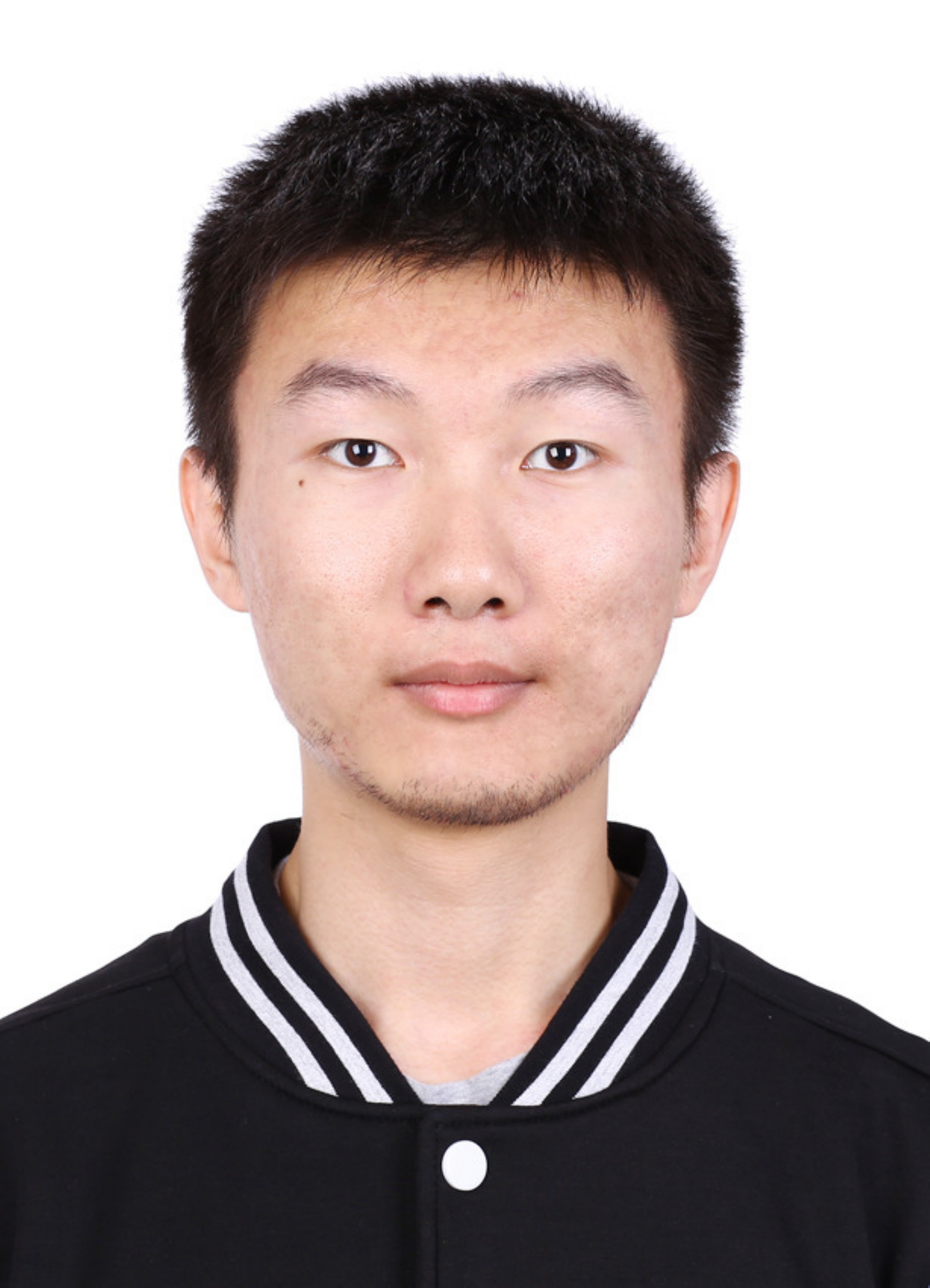}}]{Runze Liu}

received the B.S. degree in School of Computer Science and Engineering, from Beihang University, Beijing, China, in 2018.

His research interests include computing architectures for deep learning and machine vision.

\end{IEEEbiography}

\vspace{-15mm}

\begin{IEEEbiography}[{\includegraphics[width=1in,height=1.25in,clip,keepaspectratio]{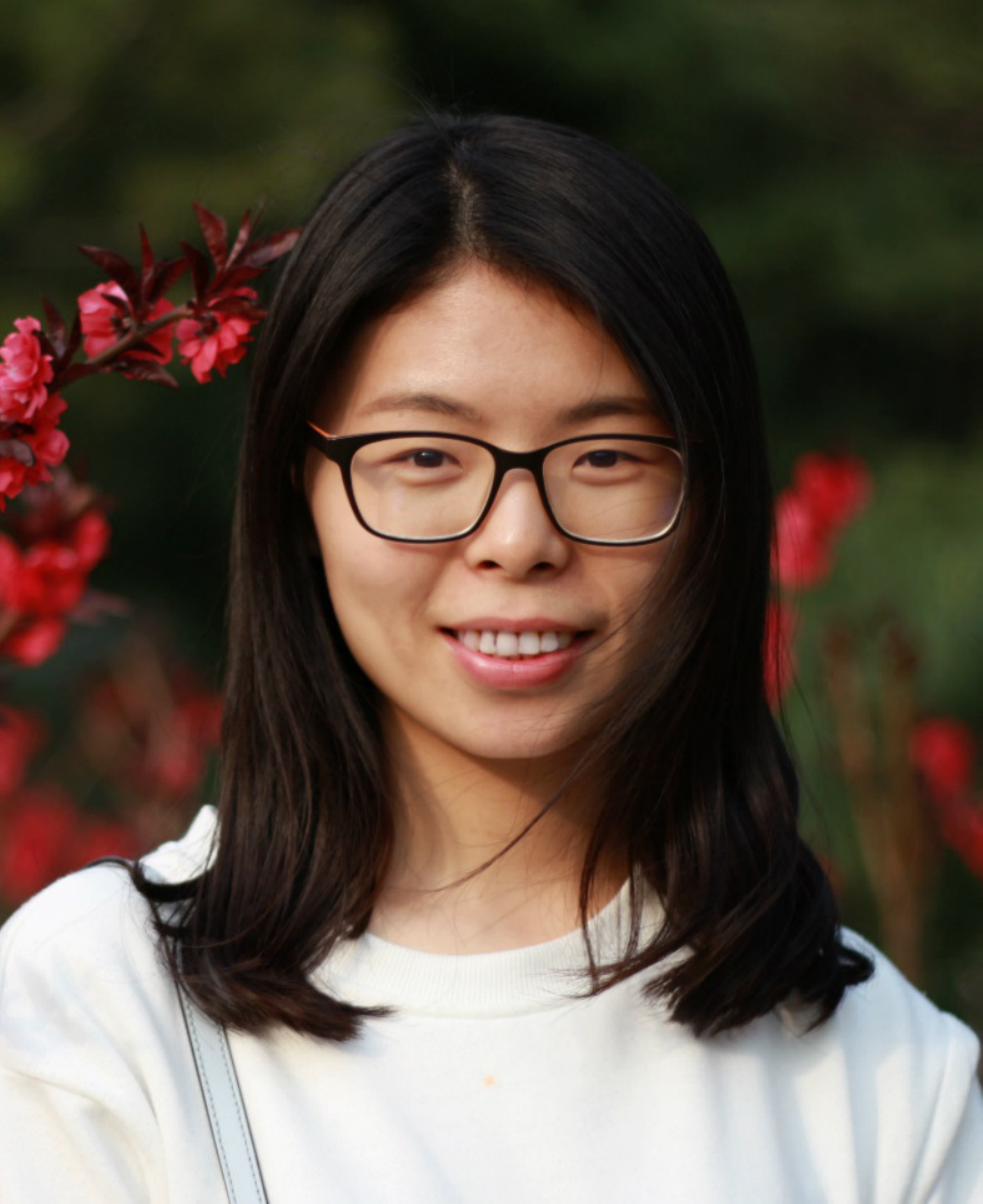}}]{Xueyan Wang}

(S'14-M'18) received the B.S. degree in computer science from Shandong University, Jinan, China, in 2013, and the Ph.D. degree in computer science and technology from Tsinghua University, Beijing, China, in 2018. From 2015 to 2016, she was a visiting scholar in University of Maryland, College Park, MD, USA.

She is currently a post-doctoral researcher with the School of Microelectronics in Beihang University, Beijing, China. Her current research interests include highly efficient processing-in-memory (PIM) architectures and hardware security.

\end{IEEEbiography}

\vspace{-15mm}

\begin{IEEEbiography}[{\includegraphics[width=1in,height=1.25in,clip,keepaspectratio]{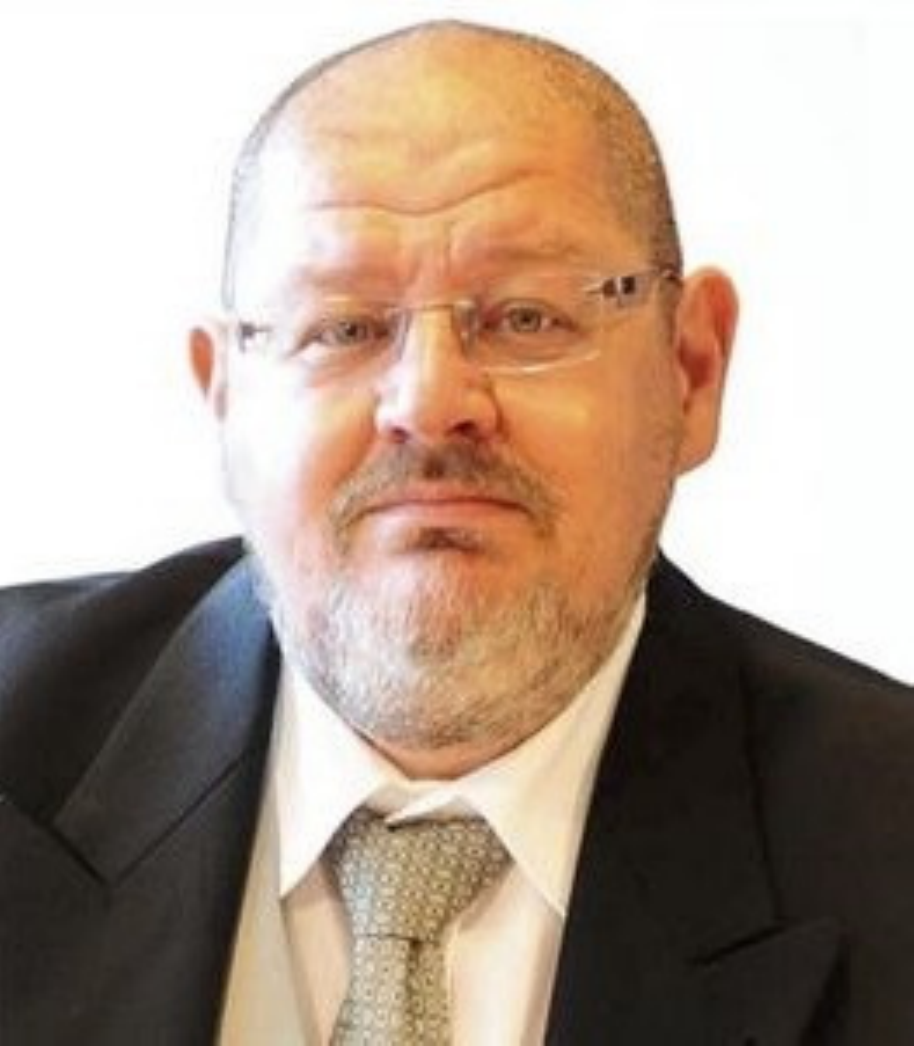}}]{Sorin Dan Cotofana}

(M'93-SM'00-F'17) received the M.Sc. degree in computer science from ``Politehnica" University of Bucharest, Romania, in 1984, and the Ph.D. degree in electrical engineering from Delft University of Technology, Delft, The Netherlands. He is currently with the Faculty of Electrical Engineering, Mathematics and Computer Science, the Computer Engineering Laboratory, Delft University of Technology, The Netherlands.

He has coauthored more than 250 papers in peer-reviewed international journal and conferences, and received 12 best paper awards in international conferences. His current research interests include the following: 1) the design and implementation of dependable/reliable systems out of unpredictable/unreliable components; 2) aging assessment/prediction and lifetime reliability aware resource management; and 3) unconventional computation paradigms and computation with emerging nano-devices.

He is currently the Editor in Chief of \textsc{IEEE Transactions on Nanotechnology}, Associate Editor for \textsc{IEEE Transactions on  Computers}, and \textsc{IEEE Circuits and Systems Society (CASS)} Distinguished Lecturer and Board of Governors member. He is an IEEE Fellow and a HiPEAC member.

\end{IEEEbiography}

\vspace{-15mm}

\begin{IEEEbiography}[{\includegraphics[width=1in,height=1.25in,clip,keepaspectratio]{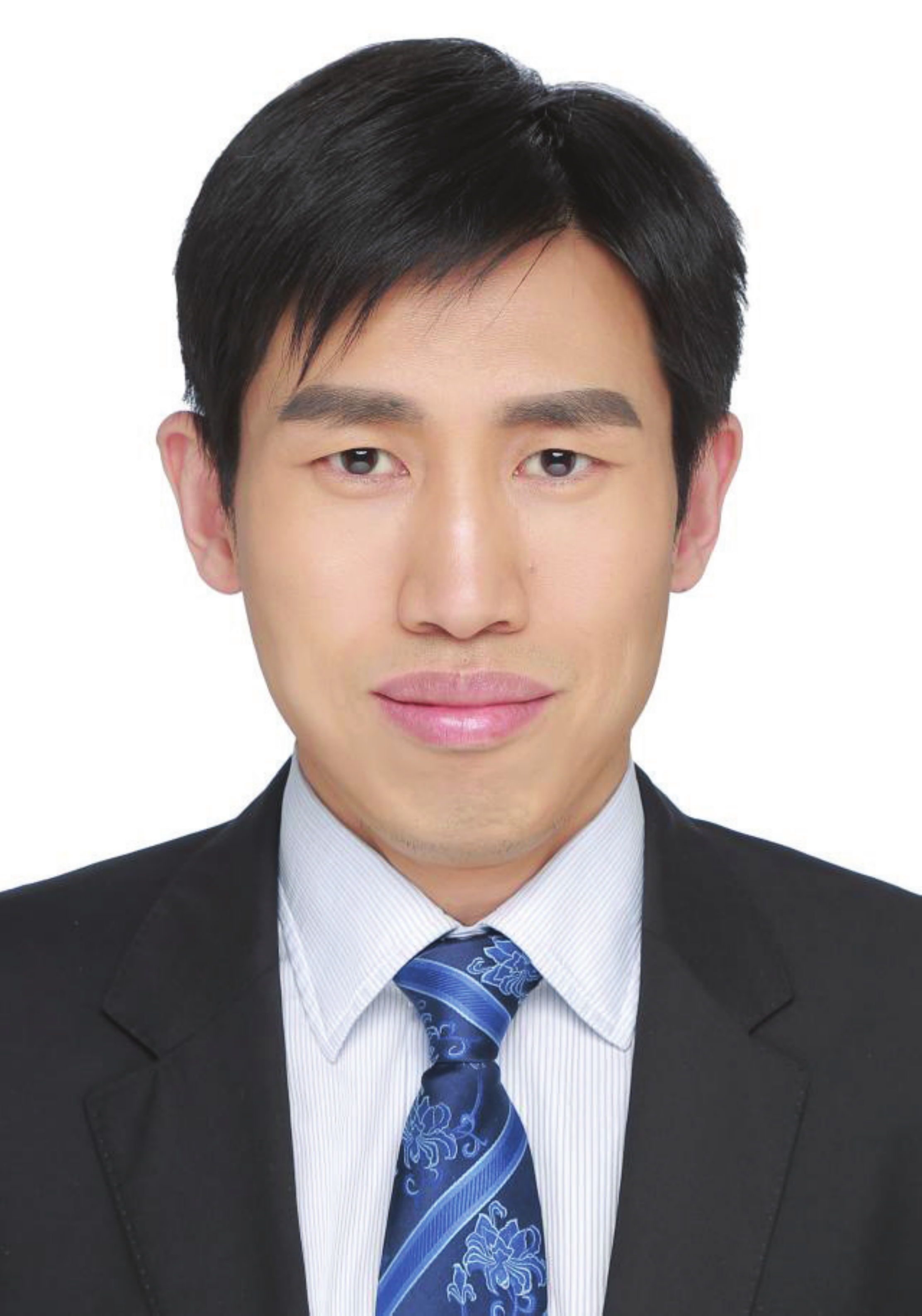}}]{Weisheng Zhao}

(M'06-SM'14-F'19) received the Ph.D. degree in physics from University of Paris Sud, Paris, France, in 2007. He is currently the Professor with the School of Microelectronics in Beihang University, Beijing, China.

In 2009, he joined the French National Research Center(CNRS), as a Tenured Research Scientist. Since 2014, he has been a Distinguished Professor with Beihang University, Beijing, China. He has published more than 200 scientific articles in leading journals and conferences, such as \textsc{Nature Electronics, Nature Communications, Advanced Materials, IEEE Transactions}, ISCA and DAC. His current research interests include the hybrid integration of nano-devices with CMOS circuit and new nonvolatile memory (40-nm technology node and below) like MRAM circuit and architecture design.

He is currently the Editor-In-Chief for the \textsc{IEEE Transactions on Circuits and Systems I: Regular Paper}. He is an IEEE Fellow.

\end{IEEEbiography}

\end{document}